# An Efficient, Probabilistically Sound Algorithm for Segmentation and Word Discovery

Michael R. Brent
Johns Hopkins University

Address correspondence to: Michael R. Brent
Department of Cognitive Science
Johns Hopkins University
Baltimore, MD 21218

email: brent@jhu.edu
phone: 410-516-6844
fax: 410-516-8020

## Abstract

This paper presents a model-based, unsupervised algorithm for recovering word boundaries in a natural-language text from which they have been deleted. The algorithm is derived from a probability model of the source that generated the text. The fundamental structure of the model is specified abstractly so that the detailed component models of phonology, word-order, and word frequency can be replaced in a modular fashion. The model yields a language-independent, prior probability distribution on all possible sequences of all possible words over a given alphabet, based on the assumption that the input was generated by concatenating words from a fixed but unknown lexicon. The model is unusual in that it treats the generation of a complete corpus, regardless of length, as a single event in the probability space. Accordingly, the algorithm does not estimate a probability distribution on words; instead, it attempts to calculate the prior probabilities of various word sequences that could underlie the observed text. Experiments on phonemic transcripts of spontaneous speech by parents to young children suggest that our algorithm is more effective than other proposed algorithms, at least when utterance boundaries are given and the text includes a substantial number of short utterances.

# 1 Introduction

## 1.1 Background and Motivation

Unlike printed English, in which words are separated by blank spaces, speech does not contain any reliable acoustic demarcation of word boundaries. As a result, the input from which children learn their native language more closely resembles a series of utterances, which *are* demarcated by silence, than a series of words. Language acquisition researchers have long been interested in how children segment speech and learn the sounds of individual words, starting from a state in which they do not know any words. This interest has led to a number of proposed algorithms for an abstract version of the speech segmentation task. In this task words are represented as strings of letters or phonemes[1] and speech is represented as a text from which the word boundaries have been removed. Such algorithms also have potential applications for segmenting written texts in languages where word boundaries are not marked in the orthography. In this paper, however, we focus on algorithms that children could use for segmentation and word discovery during language acquisition. This goal imposes significant constraints on segmentation algorithms. First, they must start out without any knowledge specific to a particular language; engineering systems can make use of existing dictionaries. Second, they must learn in a completely unsupervised fashion; engineering systems can train on pre-segmented text. Third, they must segment incrementally. To a first approximation, this means that the segmentation of each utterance must be finalized before the next utterance is read in. Specifically, algorithms that make multiple passes through the corpus or do global optimization will not be considered here.[2] Finally, the cognitive modeling goal dictates the kind of corpus on which we compare algorithms — phonemic transcripts of spontaneous speech by mothers to their young children. As discussed below, these corpora are quite different from the types of corpora normally used in language engineering; performance on child-directed speech corpora may not be a good predictor of performance on the Wall Street Journal.

Previous segmentation algorithms can be divided into three major classes. The first class focuses on identifying individual word boundaries. These algorithms do not have any explicit representations of words. They isolate words within an utterance only as a side-effect of correctly identifying two adjacent word boundaries. In a class of its own is Olivier's (1968) algorithm, which is based on maximum-likelihood estimation of a probability distribution on words and maximum-likelihood segmentation of successive blocks of input. The final class of previously proposed segmentation algorithms consists of those that are based on some form of minimum representation-length or text compression. Algorithms from each of these classes are discussed below.

The algorithms proposed in this paper are based on an explicit probability model. Given any observed input corpus, the model defines an optimization problem: Find the highest-probability segmentation of the input, where a segmentation is any sequence of words that yields the observed input when the word boundaries are deleted. Two local optimization algorithms that address this problem are discussed.

## 1.2 Boundary-finding algorithms

### 1.2.1 Local statistics

Harris (1954) sketched the first procedure for finding morpheme[3] boundaries in phonetic transcriptions of sentences. Harris defined the *successor count* of each prefix string of a sentence as the number of distinct phonemes that can follow that prefix in some grammatical sentence of the language — in other words, the number of one-phoneme extensions that are themselves prefixes of some grammatical sentence. For example, consider the utterance *he's quicker*, transcribed as /hɪyzkwɪkər/. The successor count of /hɪyzk/ is defined to be the number of distinct phonemes that can follow /hɪyzk/ in grammatical English sentences that begin with the phonemes /hɪyzk/, sentences such *he's cranky*, *he's quiet*, *he's careless*, and so on. To segment an utterance *U*, Harris proposed that boundaries be placed after those prefixes of *U* whose successor counts are at least as great as those of their immediate neighbors, the prefixes of *U* that are one phoneme shorter or longer. For example, if the successor count of /hɪ/ were 14, that of /hɪy/ were 29, and that of /hɪyzk/ were also 29, then a boundary would be placed after /hɪy/ (*he*). (See Brent & Cartwright, 1996, for further discussion.) This procedure cannot be considered an algorithm, however, since it relies on human introspection to determine the successor counts. But Harris's idea — that the successors of phonemes within words will tend to be more constrained than the successors of phonemes at the ends of words — has survived in other proposals. For example, Saffran, Newport, and Aslin (1996), treating syllables rather than phonemes as the fundamental units of input, have proposed that children might estimate the probability of each syllable in the language conditioned on its predecessor. This particular conditional probability estimate is commonly called the *transitional probability.* Saffran et al. suggest that children may segment utterances at low points of the transitional probability between adjacent syllables — that is, when a syllable occurs that is surprising given its predecessor.

Transitional probability is asymmetric — the surprisingness of each syllable is conditioned on its predecessor but not on its successor. Although it has never been proposed in the language acquisition literature, a more natural, symmetric measure of the surprisingness of a pair of adjacent sound units is their *mutual information (MI),* defined as:

$$\mathrm{MI}(x,y) \equiv \log_2 \frac{\Pr(xy)}{\Pr(x)\cdot\Pr(y)}$$

(see, e.g., Jelinek, 1997). In the experiments reported below, we compare the performance of an algorithm that segments at low points of transitional probability between phonemes, and one that segments at low points of mutual information, to our algorithm.

### 1.2.2 Connectionist algorithms

Elman (1990) proposed a connectionist segmentation algorithm based on the same idea — segment just prior to phonemes that are surprising given the preceding context. However, instead of using a statistic based on a fixed context window, such as one phoneme, Elman used a simple recurrent net (SRN) to evaluate surprisingness. An SRN is an artificial neural-network in which input is processed sequentially and the activations on the hidden layer at each time step are fed back as input to the hidden layer at the next time step. This gives the network access to a limited amount of information about the left context at each time step. The distance over which information is stored is not fixed in advance; it depends on the network's weights and on the particular input sequence. Elman encoded an alphabet of phonemes as arbitrary five-bit sequences and trained the network to predict the next phoneme at each time step. He then suggested that time steps on which the network's prediction error was high were likely to be those on which it attempted to predict the first phoneme of a word. In the experiments reported below we compare the performance of Elman's algorithm to that of our own.

Christiansen, Allen, and Seidenberg (1998) propose a segmentation algorithm that uses SRN's in a different way. Like Elman, Christiansen et al. train an SRN to predict the next symbol in the input, but they include utterance boundaries among the input symbols to be predicted. They interpret the net as predicting a *word* boundary whenever the output on the *utterance* boundary unit exceeds a certain threshold (the mean activation of the utterance boundary unit during training). The reasoning is that output on this unit will tend to rise after phoneme sequences that occur relatively often at the ends of utterances, and since the end of an utterance is also the end of a word, those same sequences will tend to occur relatively often at the ends of words (see also Aslin, Woodward, LaMendola, & Bever, 1996; Brent & Cartwright, 1996). Christiansen et al. also report simulation experiments in which the input and the prediction task included information about which phonemes were part of a stressed syllable and which were not. Their results, which are discussed further below, suggest that both the utterance-boundary strategy and the stress information are useful for segmentation, but stress is much less useful than utterance boundaries.

### 1.3 Word grammars

Olivier (1968) proposed a segmentation algorithm based on the idea of reestimating probabilistic word grammars. A word grammar is just a finite list of finite strings (i.e., a lexicon) and the language it generates is the set of all finite concatenations of those strings. A probabilistic word grammar is a word grammar along with a probability distribution on the words. Olivier's algorithm maintains an integer for each word in its grammar that corresponds in some sense to an estimate of the frequency with which that word has occurred in the input so far. Initially, each character in the alphabet is given a frequency estimate of two. (This represents an a priori estimate of the frequency with which the character occurs as a word, not as a character in other words.) The algorithm then processes the input incrementally in blocks. Before processing each block, it divides the frequency estimates of the words in its current grammar by their sum to come up with an estimated probability distribution on its current word grammar. It then finds the maximum likelihood (ML) segmentation of the block given its current word grammar, using dynamic programming. Finally, it updates the frequency counts in two ways. First, it adds to the frequency estimate of each word to that word's frequency in the ML segmentation of the current block. Second, it joins each pair of adjacent words in the ML segmentation of the current block and increments the frequency of the resulting words by one. The second procedure is necessary because Olivier's method of estimating probability distributions assigns probability zero to any word not in the current word grammar. This implies that the maximum likelihood parse can never contain any word that was not already in the word grammar. (The presence of all the letters as words in the initial grammar ensures that there is always a parse with non-zero probability.) As a result, updating the frequency estimates based only on the maximum likelihood parse would never yield any new words. In Olivier's own 1968 implementation frequency-one words are deleted from the lexicon periodically, but he explicitly states that this is not a fundamental part of the algorithm, just an expediency to allow the program to run in the available memory.

### 1.4 Information-based approaches

This section discusses approaches in which representation schemes are used to define probability distributions over segmentations implicitly. A *binary representation scheme* associates to any finite string over a given *source alphabet* one or more finite binary strings known as the *representations* of that source string, in such a way that no pair of distinct source strings shares a common representation. Alternative terms for representation scheme and representation are *code* and *encoding*, respectively. A representation scheme is said to be *self-delimiting* if there exists an algorithm that can find the end of any representation without reading beyond the end, given the beginning. For any self-delimiting binary representation scheme, the sum of the negative binary exponentials of the lengths of all representations ($\Sigma 2^{-length}$) converges to a number between zero and one (Kraft, 1949; Li & Vitanyi, 1993). This implies that the set of all source strings can be treated as a discrete probability space, where the probability of any source string is proportional to the sum of the negative binary exponentials of the lengths of all its representations. If a string has only one representation then its probability is just the negative binary exponential of that representation's length. The proportionality constant is a normalizing factor equal to one over the sum of the negative binary exponential of all representations ($1/\Sigma 2^{-length}$). Representations are useful because they provide a straightforward method of constructing a measure whose sum is guaranteed to converge.

Brent & Cartwright (1996) devised a self-delimiting representation scheme that assigns a unique representation to each *segmentation* of each string over a given alphabet. They then treated the problem of segmenting an input text as an optimization problem: Find the segmentation of the input with the shortest representation. This is equivalent to finding the most probable segmentation under the distribution in which the prior probability of each segmentation is proportional to the negative binary exponential of its representation's length. Brent and Cartwright's scheme works by Huffman-coding the sequence of words in a segmented text, then coding the mapping from Huffman codes to words (see Figure 1 for an example).

**Insert Figure 1 about here.**

The paper reported on the segmentation accuracy achieved when a particular heuristic strategy was used to search for the segmentation of a corpus that minimized this objective function. However, this search strategy was off-line and took time proportional to the cube of the number of phonemes in the corpus. Since this search mechanism was not cognitively plausible, Brent and Cartwright limited their cognitive model to the objective function. The relation of that function to the probability model proposed here is discussed below, as are the results Brent and Cartwright reported.

Brent and Cartwright's approach to segmentation is one example of the minimum representation-length technique, also known as minimum message-length (Wallace & Bolton, 1968) and minimum description-length (Quinlan & Rivest, 1989; Rissanen, 1989). It is traditional in the minimum representation-length literature to divide representations into two portions, one corresponding to the set of generalizations extracted from the input (lexicon, grammar, decision tree, patterns, rules) and the other corresponding to the "accidental" or "unpredictable" component of the input, such as what a particular person chose to say at a particular time, given the constraints imposed by the rules of his or her language. The length of each portion can then be interpreted in Bayesian terms as the probability of the generalizations and the probability of the unpredictable component, given the generalizations. This division often leads to useful insights about how to interpret the shortest representation of an input in terms of some other inference problem. Dividing representations up in this way is not necessary for the construction of a probability distribution, however.

Text compression schemes like LZW (the basis of *compress*, see e.g. Hankerson, Harris, and Johnson, 1998) are also self-delimiting representation schemes for texts over a source alphabet. Further, they represent source texts by assigning them a unique segmentation into shorter strings and applying a self-delimiting representation scheme to each shorter string in the resulting sequence. The compressed text can be viewed as a representation of the unique segmentation that LZW assigns to the source text. Because it assigns a unique segmentation to each source string, however, it is not a representation scheme for all possible segmentations of all source strings. Viewed as a probability distribution on segmentations of a given input, LZW assigns all the probability to one segmentation. The segmentations chosen by a deterministic scheme like LZW could turn out to have linguistic relevance, but there is no a priori reason to expect that they should. We return to the linguistic relevance of LZW in the experimental section.

De Marcken (1996) also took a minimum representation-length approach to the segmentation problem. However, rather than representing segmentations of the source string, de Marcken's scheme represents a hierarchical decomposition where the input is divided into substrings and each of the substrings is further subdivided until the level of individual characters is reached (see also Nevill-Manning & Witten, 1997; Wolff, 1982). This structure can be viewed as a parse tree whose leaves are the individual characters of the input, and whose nodes span the substrings of the decomposition. Given a text, de Marcken used an off-line algorithm to search for the hierarchical decomposition with the shortest representation. At the end, the substring spanned by each node in each tree was interpreted as a possible word, but there was no commitment about which actually were words and which were not. Thus, while de Marcken's algorithm and hierarchical text compression algorithms are related to the problem of interest here, they do not address that problem directly.

Redlich (1993) also proposed an off-line segmentation algorithm based loosely on the minimum representation-length framework. However, the only experimental results reported were on one paragraph of written text. Further, the algorithm was not sufficiently well specified that we could reimplement it, and hence it is difficult to evaluate.

### 1.4.1 Information minimization versus explicit probability models

The minimum representation-length approach provides a straightforward, intuitive method for constructing a probability distribution. Further, it is not difficult to craft a representation scheme such that the corresponding distribution roughly reflects one's intuitions about the structure to be uncovered — that is, the system responsible for generating the observed text. However, the representation schemes one devises are very rarely perfectly efficient; that is, there are binary strings *b* such that *b* is not a prefix of the representation of any source string, nor is the representation of any source string a prefix of *b*. This implies that the sum over all source strings of the corresponding probability is strictly less than one — there is missing probability mass and there may not be any sensible model that generates source strings according to the induced measure. There is no way to know how this missing mass may influence the inference process without recovering it or characterizing it formally. Even when the missing mass is relatively small in practical terms, there are theoretical advantages to having a coherent probabilistic model of how the input is generated. A sound model of the source yields insight into how the model corresponds to known linguistic structure, how it differs, and how the differences might be mitigated.

Brent and Cartwright's representation scheme offers an example of missing probability. It chooses one particular assignment of binary codes to words even though any consistent assignment could represent the same segmentation. All the potential representations based on alternative assignments are wasted — they don't represent any segmentation, they are not prefixes of the representation of any segmentation, and they are not extensions of the representation of any segmentation. As a result, the probability mass corresponding to these potential representations is lost.

In the next section of this paper we present an explicit probabilistic model of how lexicons are generated from alphabets of phonemes and how texts are generated by concatenating words from lexicons. The general structure of the resulting probability measure is similar to that which Brent and Cartwright (1996) derived by the minimum representation-length method. However, some of the missing probability mass has been recovered, and all implicit approximations have been uncovered and made explicit. The explicit probability model also makes it clear how components of the model, such as the distribution on the phonological forms of words, can be upgraded in a modular fashion. Indeed, one small step in that direction is taken in this paper, where we estimate a distribution on phonemes in the lexicon rather than using a uniform distribution.

## 2 A probability distribution on all finite sequences of all possible words

This section introduces an explicit, language-independent, model-based probability distribution on all possible sequences of all possible words over a given alphabet. The presentation is organized into four subsections. The first subsection describes the structure of the probability model at a level that abstracts away from many details needed to actually compute the probabilities of word sequences. The abstract structure is quite general and independent of the various simplifying assumptions that are made later on for the sake of expediency; it serves as a backbone to which various component models can be attached.

The second subsection derives a pair of recursive formulae for evaluating the probability of any given word sequence. These formulae are stated in terms of the abstract model; they cannot be used for computing probabilities until the detailed component models are specified.

The third section presents one set of component models that are motivated, in part, by algebraic and computational simplicity. It is expected that each one will be improved in future work. This subsection concludes with the specific recursive formulae used in the experiments reported below.

The fourth and final subsection discusses the equations derived in the previous three sections. The discussion focuses on the implications of an interesting theoretical observation about the probability model: The prior probabilities of segmentations can be evaluated without estimating a probability distribution on words. In specifying the detailed distributions we assume that a distribution on phonemes is estimated from the input, but that is not necessary. If a uniform distribution on phonemes were used instead then the probability of any given word sequence could be evaluated exactly, without estimating anything.

It is worth noting that the goals of this section are primary theoretical. They are: (a) to provide a general purpose framework to which a variety of specific linguistic distributions can be interfaced in a modular fashion, and (b) to show one example of how this can be done. Naturally, the probability model presented in this section defines an optimization problem: Given an unsegmented input, find the most probable word sequence among all possible segmentations of the input. Two incremental algorithms aimed at finding an approximate solution to this problem quickly are introduced in the next section. In the current section the focus is on the model and on the evaluation of the prior probabilities of individual segmentations.

### 2.1 Structure of the model

This subsection describes the structure of the probability model at a level that abstracts away from many details needed to actually compute the probabilities of word sequences. The structure is specified as a non-deterministic algorithm that generates every possible word sequence and then, by deleting word boundaries, every possible input to a segmentation procedure.

Let $\Sigma$ be the input alphabet and let # and $ be two symbols not contained in $\Sigma$. The generation algorithm consists of four non-deterministic steps and one deterministic step. At the end of the non-deterministic steps, the algorithm has generated a sequence of utterances separated by $. Each utterance consists of a sequence of words separated by #. This will be called a *delimited utterance sequence.* The final, deterministic step of the algorithm deletes the #'s leaving a sequence of utterances in which word boundaries are no longer marked. This will be called a *non-delimited utterance sequence*. Given a particular delimited utterance sequence, the result of deleting the #'s will be called its *yield*. Conversely, given a particular non-delimited sequence *N*, each delimited sequence of which *N* is the yield will be called a *segmentation* of *N*.

The input to a segmentation algorithm is a non-delimited utterance sequence; the output is a delimited utterance sequence that is a segmentation of the input.

The five steps of the abstract generative model are as follows:

1. Pick a positive integer *n*, representing the number of distinct word types to be generated.

2. Pick a set of *n* distinct strings from the set $\Sigma^{+}\#$, representing the phonological forms of word types. Call the result *L,* for lexicon. Let the names $W_1,\ldots,W_n$ be assigned arbitrarily to the members of *L*. We can now write $L=\{W_1,\ldots,W_n\}$. Let $W_0 = \$$, the distinguished utterance-boundary marker.

3. Pick a function $\mathbf{f}:\{0,\ldots,n\}\rightarrow\{1,2,\ldots\}$, where $\mathbf{f}(i)$ represents the total frequency of word $W_i$. Note that f($i$) is not a relative frequency or probability but a positive integer representing the number of times word $W_i$ will occur in the word sequence being generated.

4. Let *m* be the total number of word tokens — the sum of the frequencies of all *n* words. Pick an ordering function $\mathbf{s}:\{1,\ldots,m\}\rightarrow\{1,\ldots,n\}$ that maps each position in the text to be generated to the index of the word that will appear in that position, so that $W_{\mathbf{s}(i)}$ appears as the $i^{\text{th}}$ word token in the generated text. For notational convenience, define

$$w_1,\ldots,w_m \equiv W_{s(1)},\ldots,W_{s(m)}.$$

Note that $w_1,\ldots,w_m$ is a delimited utterance sequence. Define $\overline{w_m} \equiv w_1,\ldots,w_m$. Let the domain of **s** be called the set of token indices and the range of **s** be called the set of type indices. Note that **s** is constrained to map exactly $\mathbf{f}(i)$ token positions onto type index *i*.

5. Concatenate $w_1,\ldots,w_m$, delete the #'s, and output the result. The output is a non-delimited utterance sequence, the yield of $w_1,\ldots,w_m$.

As a concrete example the generative process could yield the results shown in Table 1 on some run, using ordinary letters for the alphabet Σ.

**Insert Table 1 about here**

In order to evaluate a particular hypothesized segmentation $\overline{w}_m = w_1, \ldots, w_m$ (including word and utterance-boundary markers) we need a formula for the probability with which that word sequence is generated by steps 1-4 of the model, $\Pr(\overline{w}_m)$. Starting from first principles, this means summing over all possible outcomes of steps 1, 2, 3, and 4:

$$\Pr(\overline{w}_m) = \sum_n \sum_L \sum_{\mathrm{f}} \sum_s \Pr(\overline{w}_m | n, L, \mathrm{f}, \mathrm{s}) \cdot \Pr(n, L, \mathrm{f}, \mathrm{s})$$

where each sum is over all possible values of the corresponding variable. However, the abstract model specified above was designed so that there is a one-to-one correspondence between hypothesized word sequences and joint outcomes of steps 1-4. Specifically, for any given hypothesis $\overline{w}_m$, there is one and only one combination of values of $n$, $L$, **f**, and **s** for which $\Pr(\overline{w}_m | n, L, \mathrm{f}, \mathrm{s}) \neq 0$ — namely, $n$ must be the number of distinct word types in $\overline{w}_m$, $L$ must be the set of distinct word types in $\overline{w}_m$ (since the frequencies are strictly positive), **f** must map each word type in $L$ into the number of times it appears in $\overline{w}_m$, and **s** must be such that $W_{\mathbf{s}(1)}, \ldots, W_{\mathbf{s}(m)} = w_1, \ldots, w_m$. Since these particular values of $n$, $L$, **f**, and **s** are completely determined by $\overline{w}_m$, it would make sense to write them as functions of $\overline{w}_m$ — for example, we could write the number of distinct word types in $\overline{w}_m$ as $\mathrm{n}(\overline{w}_m)$. However, this notation is bulky, so the notation $n_m$, $L_m$, $\mathbf{f}_m$, and $\mathbf{s}_m$ is used instead. Since there is only one non-zero term in the sum, we can write:

$$\Pr(\overline{w}_m) = \Pr(\overline{w}_m | n_m, L_m, \mathrm{f}_m, \mathrm{s}_m) \cdot \Pr(n_m, L_m, \mathrm{f}_m, s_m)$$

Furthermore, since the values of $n$, $L$, **f**, and **s** completely determine $\overline{w}_m$,

$$\Pr(\overline{w}_m | n_m, L_m, \mathrm{f}_m, \mathrm{s}_m) = 1$$

and we have:

$$\Pr(\overline{w}_m) = \Pr(n_m, L_m, \mathrm{f}_m, s_m)$$

Using the chain rule, this can be rewritten as:

(1) $$\Pr(\overline{w}_m) = \Pr(s_m|\mathrm{f}_m, L_m, n_m)\cdot\Pr(\mathrm{f}_m|L_m, n_m)\cdot\Pr(L_m|n_m)\cdot\Pr(n_m)$$

Now we add to the abstract model two reasonable linguistic assumptions that allow this expression to be simplified substantially.

1. The ordering of the words, **s**, is probabilistically independent of their pronunciations, so $\Pr(\mathrm{s}_m|\mathrm{f}_m, L_m, n_m) = \Pr(\mathrm{s}_m|\mathrm{f}_m, n_m)$. This does not necessarily mean that the ordering is independent of all aspects of word identity. If we had modeled other properties of words, such as their syntactic categories, the ordering might be dependent on those properties. But the assumption implies that any effect of pronunciation on ordering would be mediated by other properties. This is not true in song lyrics or poetry, where word combinations are chosen partly on the basis of their sound, but it is true to a first approximation in other forms of language.

    The frequency function $\mathrm{f}_m$ completely determines the number of distinct words $n_m$, since the domain of $\mathrm{f}_m$ consists of $n_m$ integers, so $\Pr(\mathrm{s}_m|\mathrm{f}_m, n_m) = \Pr(\mathrm{s}_m|\mathrm{f}_m)$. (This is just a fact about the way $\mathrm{f}_m$ and $n_m$ were defined.)

2. The frequencies of words are chosen independently of the frequencies and pronunciations of all other words, and of $n$, so $\Pr(\mathrm{f}_m|L_m, n_m) = \prod_{i=0}^{n_m}\Pr_{\mathrm{f}}(\mathrm{f}(i)|W_i)$, where $\Pr_{\mathrm{f}}$ is the probability distribution on the frequencies of individual words. Note that this does not imply a uniform distribution on word frequencies — like outcomes on the sum of two dice, word frequencies can have a nonuniform distribution despite being selected independently of one another.

    Since $\mathrm{f}_m(0)$ represents the number of utterance-boundary markers, we are also assuming by this equation that the frequencies of words are independent of the number of

utterances. The latter assumption is not accurate — if some word occurs 1,000 times, it is unlikely that the total number of utterances is as small as three. However, this inaccuracy has little importance, since the actual number of utterances is observable in the input and hence identical for all segmentations of the input.

These two assumptions allow us to simplify (1) to:

$$\text{(2)}\qquad \Pr(\overline{w}_m) = \Pr(\mathrm{s}_m|\mathrm{f}_m)\cdot\left[\prod_{i=0}^{n_m}\Pr_{\mathrm{f}}(\mathrm{f}(i)|W_i)\right]\cdot\Pr(L_m|n_m)\cdot\Pr(n_m)$$

In order to evaluate this we need to make further assumptions about how each of the non-deterministic steps of the generative process works. Before moving on to such assumptions, however, we introduce recursive formulae for evaluating (2).

## 2.2 Relative probabilities

This section introduces two recursive formulae for evaluating the probability of a word sequence $w_1, \ldots, w_k$ in terms of the probability of the sequence $w_1, \ldots, w_{k-1}$. Let the notation R (for *relative probability*) be defined as follows:[4]

$$\text{(3)}\qquad \mathrm{R}(\overline{w}_k) \equiv \frac{\Pr(\overline{w}_k)}{\Pr(\overline{w}_{k-1})},$$

where $\overline{w}_k \equiv w_1, \ldots, w_k$, and $\Pr(\overline{w}_0) \equiv 1$. Observe that:

$$\text{(4)}\qquad \Pr(\overline{w}_k) = \mathrm{R}(\overline{w}_k)\cdot\Pr(\overline{w}_{k-1}),$$

a recursive formula, and:

(5) $$\Pr(\overline{w}_k) = \prod_{i=1}^{k} \mathrm{R}(\overline{w}_i)$$

Hereafter, we focus on computing $\mathrm{R}(\overline{w}_k)$. Note that $\mathrm{R}(\overline{w}_k)$ is undefined if $\overline{w}_{k-1}$ has probability zero.

Now for any given $\overline{w}_k$, we can define $n_k$, $L_k$, $\mathbf{f}_k$, and $\mathbf{s}_k$. analogously to $n_m$, $L_m$, $\mathbf{f}_m$, and $\mathbf{s}_m$, as the number of distinct words, set of distinct words, word frequencies, and ordering of words in $\overline{w}_k$, respectively. Whenever R is defined, we can substitute (2) into (3), yielding:

(6) $$\mathrm{R}(\overline{w}_k) = \frac{\Pr(\mathrm{s}_k|\mathrm{f}_k)}{\Pr(\mathrm{s}_{k-1}|\mathrm{f}_{k-1})} \cdot \frac{\prod_{i=0}^{n_k} \Pr_{\mathrm{f}}(\mathrm{f}_k(i)|W_i)}{\prod_{i=0}^{n_{k-1}} \Pr_{\mathrm{f}}(\mathrm{f}_{k-1}(i)|W_i)} \cdot \frac{\Pr(L_k|n_k)}{\Pr(L_{k-1}|n_{k-1})} \times \frac{\Pr(n_k)}{\Pr(n_{k-1})}$$

This formula can be simplified by separating two cases, one in which the *k*th word also occurs in the first k-1 words — that is, $w_k \in L_{k-1}$ — and one in which it does not.

### 2.2.1 Case 1: Familiar words

Word $w_k$ is called a familiar word if it also occurs in the first *k*-1 words — that is, $w_k \in L_{k-1}$. In that case, adding $w_k$ onto the end of $w_1, \ldots, w_{k-1}$ leaves the number of distinct words and the set of distinct words unchanged ($n_k = n_{k-1}$, $L_k = L_{k-1}$), so

$$\mathrm{R}(\overline{w}_k | w_k \in L_{k-1}) = \frac{\Pr(\mathrm{s}_k|\mathrm{f}_k)}{\Pr(\mathrm{s}_{k-1}|\mathrm{f}_{k-1})} \cdot \frac{\prod_{i=0}^{n_k} \Pr_{\mathrm{f}}(\mathrm{f}_k(i)|W_i)}{\prod_{i=0}^{n_k} \Pr_{\mathrm{f}}(\mathrm{f}_{k-1}(i)|W_i)},$$

where $R(\overline{w}_k \mid w_k \in L_{k-1})$ denotes the relative probability given that the $k$th word also occurs among the first $k-1$ words. Further, adding $w_k$ onto the end of $w_1, \ldots, w_{k-1}$ increases the frequency of $w_k$ by one ($\mathbf{f}_k(i) = \mathbf{f}_{k-1}(i)+1$, for $i = \mathbf{s}_k(k)$ — that is, when $i$ is the index for the word type corresponding to $w_k$) but leaves the frequencies of all other words unchanged ($\mathbf{f}_k(i) = \mathbf{f}_{k-1}(i)$, for $i \neq \mathbf{s}_k(k)$). If we define $\hat{k}$ to be the type index of the $k$th word ($\hat{k} \equiv \mathrm{s}_k(k)$) then we have $\mathrm{f}_k(\hat{k}) = \mathrm{f}_{k-1}(\hat{k}) + 1$ and hence:

$$(7) \qquad R(\overline{w}_k \mid w_k \in L_{k-1}) = \frac{\Pr(\mathrm{s}_k \mid \mathrm{f}_k)}{\Pr(\mathrm{s}_{k-1} \mid \mathrm{f}_{k-1})} \cdot \frac{\Pr_{\mathrm{f}}(\mathrm{f}_k(\hat{k}) \mid W_{\hat{k}})}{\Pr_{\mathrm{f}}(\mathrm{f}_k(\hat{k}) - 1 \mid W_{\hat{k}})}$$

This expression cannot be simplified any more without making further assumptions about the distribution on ordering functions. To see the general form that such simplifications might take, note that $\mathbf{s}_k(i) = \mathbf{s}_{k-1}(i)$ for all $i < k$. Thus, any distribution on orderings that has some locality will yield related probabilities for the two orderings $\mathbf{s}_k$ and $\mathbf{s}_{k-1}$.

#### 2.2.2 Case 2: Novel words

Word $w_k$ is called a novel word if its occurrence at position $k$ is its first occurrence — that is, $w_k \notin L_{k-1}$. In that case, the number of distinct words increases by one ($n_k = n_{k-1}+1$), the frequencies of the familiar words are unchanged ($\mathbf{f}_k(i) = \mathbf{f}_{k-1}(i)$ for $i < n_k$), and the frequency of the new word is one ($\mathbf{f}_k(n_k) = 1$) so (6) can be simplified to:

$$(8) \; R(\overline{w}_k \mid w_k \notin L_{k-1}) = \frac{\Pr(\mathrm{s}_k \mid \mathrm{f}_k)}{\Pr(\mathrm{s}_{k-1} \mid \mathrm{f}_{k-1})} \cdot \Pr_{\mathrm{f}}(1 \mid W_{n_k}) \cdot \frac{\Pr(L_k \mid n_k)}{\Pr(L_{k-1} \mid n_k - 1)} \cdot \frac{\Pr(n_k)}{\Pr(n_k - 1)}$$

This expression cannot be further simplified without making assumptions about the detailed distributions that govern the steps of the abstract model.

Before new assumptions are introduced in the next section, it is worth emphasizing that no new assumptions were introduced during the derivation of equations (7) and (8). In particular, evaluating the probability of a word sequence via equations (7), (8), and (5), is mathematically equivalent to using equation (2) directly.

### 2.3 Modular linguistic models

This subsection presents one among many possible ways of filling in the details of the probability model. The particular assumptions made here are approximations motivated by simplicity — it is expected that future work will improve on them. Since the assumptions are related to one another through the equations derived above, it should be possible to modify them independently. In the following presentation, each subsection corresponds to one distributional assumption.

#### 2.3.1 Uniform distribution on word orders (model step 4)

In this paper, we make the simplifying assumption that the distribution on ordering functions given frequencies ($\mathbf{Pr}(\mathbf{s}_k \,|\, \mathbf{f}_k)$) is uniform. This is equivalent to ignoring the constraints that syntax and semantics impose on word order. The main virtues of this assumption are algebraic and computational simplicity. However, there may well be efficient ways of using bigram, trigram, or more complex distributions within the same abstract model.

Consider the multiset consisting of $\mathbf{f}_k(0)$ copies of $W_0$, $\mathbf{f}_k(1)$ copies of $W_1$, and so on, up to $\mathbf{f}_k(n_k)$ copies of $W_{n_k}$. The number of distinct permutations of this multiset is:

$$\frac{k!}{\prod_{i=0}^{n_k} \mathrm{f}_k(i)!}$$

where $k$ is the sum of the frequencies of all the words (see any introduction to discrete probability). So, the probability of any particular permutation under the uniform distribution on all distinct permutations is given by:

$$\Pr_U(\mathrm{s}_k|\mathrm{f}_k) = \frac{\prod_{i=0}^{n_k} \mathrm{f}_k(i)!}{k!}$$

and

(9) $$\frac{\Pr_U(\mathrm{s}_k|\mathrm{f}_k)}{\Pr_U(\mathrm{s}_{k-1}|\mathrm{f}_{k-1})} = \frac{\prod_{i=0}^{n_k} \mathrm{f}_k(i)!}{\prod_{i=0}^{n_{k-1}} \mathrm{f}_{k-1}(i)!} \cdot \frac{(k-1)!}{k!}$$

Recall that for familiar words, the frequency of the last word increases by one ($\mathrm{f}_k(\hat{k}) = \mathrm{f}_{k-1}(\hat{k}) + 1$) and the frequencies of all other words are unchanged ($\mathrm{f}_k(i) = \mathrm{f}_{k-1}(i),\ \text{for } i \neq \hat{k}$), where $\hat{k}$ is the type index of the familiar word. Thus,(9) can be simplified to:

(10) $$\frac{\Pr_U(\mathrm{s}_k|\mathrm{f}_k)}{\Pr_U(\mathrm{s}_{k-1}|\mathrm{f}_{k-1})} = \frac{\mathrm{f}_k(\hat{k})}{k}$$

Substituting this back into (7), the equation for the relative probability of a familiar word, yields:

(11) $$\mathrm{R}(\overline{w}_k|w_k \in L_{k-1}) = \frac{\mathrm{f}_k(\hat{k})}{k} \cdot \frac{\Pr_{\mathrm{f}}(\mathrm{f}_k(\hat{k})|W_{\hat{k}})}{\Pr_{\mathrm{f}}(\mathrm{f}_k(\hat{k})-1|W_{\hat{k}})}$$

When $w_k$ is a novel word ($w_k \notin L_{k-1}$) the number of distinct words increases by one ($n_k=n_{k-1}+1$), the frequencies of the familiar words are unchanged ($\mathbf{f}_k(i) = \mathbf{f}_{k-1}(i)$ for $i < n_k$), and the frequency of the new word is one ($\mathbf{f}_k(n_k)= 1$), so (9) can be simplified to:

(12) $$\frac{\Pr_U(s_k|f_k)}{\Pr_U(s_{k-1}|f_{k-1})} = \frac{1}{k}$$

Substituting this back into (8), the equation for the relative probability of a novel word, yields:

(13) $$R(\overline{w}_k|w_k \notin L_{k-1}) = \frac{\Pr_f(1|W_{n_k})}{k} \cdot \frac{\Pr(L_k|n_k)}{\Pr(L_{k-1}|n_k - 1)} \cdot \frac{\Pr(n_k)}{\Pr(n_k - 1)}$$

### 2.3.2 Independence of word frequencies from word pronunciations (model step 3)

For purposes of this paper it is assumed that the distribution on word frequencies is independent of the pronunciations of the words. This is not strictly true — shorter words tend to have relatively higher frequency, and modeling such dependencies might well be a fruitful avenue to pursue, but it is beyond the scope of this paper. Under this assumption (11), the equation for the relative probability of a familiar word, can be rewritten as:

(14) $$R(\overline{w}_k|w_k \in L_{k-1}) = \frac{f_k(\hat{k})}{k} \cdot \frac{\Pr_f(f_k(\hat{k}))}{\Pr_f(f_k(\hat{k}) - 1)}$$

where $\mathbf{Pr}_f$ is the probability distribution on the integers used for picking word frequencies. Likewise (13), the equation for the relative probability of a novel word, can be rewritten as:

(15) $$R(\overline{w}_k|w_k \notin L_{k-1}) = \frac{\Pr_f(1)}{k} \cdot \frac{\Pr(L_k|n_k)}{\Pr(L_{k-1}|n_k - 1)} \cdot \frac{\Pr(n_k)}{\Pr(n_k - 1)}$$

### 2.3.3 Distribution on sets of pronunciations (model step 2)

The most difficult aspect of this model is finding a distribution on sets of pronunciations that is both natural from a linguistic point of view and efficiently evaluable. In fact, we have not been able to satisfy both constraints completely. We start with a linguistically natural distribution but

use an approximation to speed evaluation.

Consider a procedure that goes through $n_k$ iterations. At the $i$th iteration, it picks a string $S_i \in \Sigma^+\#$ according to some distribution $\mathbf{Pr}_\sigma$ on $\Sigma^+\#$, with the restriction that $S_i$ must be distinct from all strings picked on previous iterations. Let us define $\Pr(\overline{W}|n_k)$ to be the probability that this process produces the sequence of distinct strings $(W_1,\ldots,W_{n_k})$, in that particular order. Then

$$\begin{aligned}\Pr(\overline{W}|n_k) &= \prod_{i=1}^{n_k} \Pr_\sigma(S_i = W_i | S_i \notin \{W_1,\ldots,W_{n_k}\}) \\ &= \prod_{i=1}^{n_k} \frac{\Pr_\sigma(S_i = W_i)}{1-\Pr_\sigma(S_i \in \{W_1,\ldots,W_{n_k}\})}\end{aligned}$$

Note that the probability expression in the denominator is not the probability of drawing a particular lexicon but simply the probability that a single string selected according to $\mathbf{Pr}_\sigma$ is in a particular set. Since $\mathbf{Pr}_\sigma$ is a discrete probability distribution, the probability of a set of strings is just the sum of the probabilities of the strings in the set. Dropping the $S_i$, we can write:

$$\Pr(\overline{W}|n_k) = \prod_{i=1}^{n_k} \frac{\Pr_\sigma(W_i)}{1-\sum_{j=1}^{i-1}\Pr_\sigma(W_j)}$$

The probability with which this process produces the set of strings $L_k$ in any order is the sum, over all permutations $(W_{p(1)},\ldots,W_{p(n_k)})$ of $(W_1,\ldots,W_{n_k})$, of the probability with which it produces the sequence $(W_{p(1)},\ldots,W_{p(n_k)})$.

$$\Pr(L_k|n_k) = \sum_{p:\{1,\ldots,n_k\}} \prod_{i=1}^{n_k} \frac{\Pr_\sigma(W_{p(i)})}{1-\sum_{j=1}^{i-1}\Pr_\sigma(W_{p(j)})}$$

where the outer sum is over all permutation functions $p$ on the integers from 1 to $n_k$. The terms in the product of numerators are the same regardless of the permutation, and since multiplication is

order-independent, we can factor the product of numerators out, yielding:

$$\Pr(L_k|n_k) = \left[\prod_{i=1}^{n_k} \Pr_\sigma(W_i)\right] \sum_{p:\{1,\ldots,n_k\}} \prod_{i=1}^{n_k} \frac{1}{1-\sum_{j=1}^{i-1} \Pr_\sigma(W_{p(j)})}$$

Turning now to relative probabilities, we have:

$$\text{(16)} \qquad \frac{\Pr(L_k|n_k)}{\Pr(L_{k-1}|n_k-1)} = \Pr_\sigma(W_{n_k}) \frac{\sum_{\mathrm{p}:\{1,\ldots,n_k\}} \prod_{i=1}^{n_k} \left(1-\sum_{j=1}^{i-1} \Pr_\sigma(W_{\mathrm{p}(j)})\right)^{-1}}{\sum_{\mathrm{p}:\{1,\ldots,n_k-1\}} \prod_{i=1}^{n_k-1} \left(1-\sum_{j=1}^{i-1} \Pr_\sigma(W_{\mathrm{p}(j)})\right)^{-1}}$$

We do not know of any way to evaluate this expression exactly without summing explicitly over all permutations. This would require time exponential in the total size of the input, so we use the following approximation, which can be evaluated more efficiently:

$$\text{(17)} \qquad \frac{\Pr(L_k|n_k)}{\Pr(L_{k-1}|n_k-1)} \approx \frac{n_k \Pr_\sigma(W_{n_k})}{1-\frac{n_k-1}{n_k}\cdot\sum_{j=1}^{n_k-1} \Pr_\sigma(W_{\mathrm{p}(j)})}$$

The reasoning behind this approximation is somewhat involved and off the main track, so it has been relegated to Appendix A. Substituting (17) into (15) yields:

$$\text{(18)} \qquad \mathrm{R}(\overline{w}_k|w_k \notin L_{k-1}) = \frac{\Pr_{\mathrm{f}}(1)}{k} \cdot \frac{n_k \Pr_\sigma(W_{n_k})}{1-\frac{n_k-1}{n_k}\cdot\sum_{j=1}^{n_k} \Pr_\sigma(W_j)} \cdot \frac{\Pr(n_k)}{\Pr(n_k-1)}$$

Now all that remains is to choose a distribution $\mathbf{Pr}_\sigma$ on the space of possible pronunciations ($\Sigma^+\#$) and distributions on the positive integers.

### 2.3.4 Distribution on individual pronunciations (model step 2)

In general, languages impose a rich array of phonological constraints on their words. For

example, languages impose restrictions on the consonant clusters that can occur at the beginnings and ends of words. Ideally, a model for the probability that a particular phoneme string is a word in a particular unknown language should include a universal catalogue of all such phonological constraints and processes, appropriately parameterized to account for cross-linguistic variation. In this first attempt to spell out the details of the segmentation model, however, we sidestep phonology and phonotactics entirely and assume that the phonemes in a word are selected independently of one another.

Let $\mathbf{Pr}_\Sigma$ be a probability distribution on $\Sigma \cup \{\#\}$. Then if $a_1,\ldots,a_q \in \Sigma^+\#$ we define

(19) $$\Pr_\sigma(a_1 \ldots a_q) \equiv \frac{1}{1-\Pr_\Sigma(\#)} \cdot \prod_{i=1}^{q} \Pr_\Sigma(a_i)$$

The first term results from imposing the condition that the empty word "#" cannot be in the lexicon. This definition of $\Pr_\sigma$ induces a distribution in which the probabilities of words are bounded above by an exponentially decreasing function of word length. It is likely that the true distribution on lengths of word types has a mode somewhere between 3 and 5 phonemes, rather than 1, so more accurate models could almost certainly be found.

In the experiments reported below, we estimate $\mathbf{Pr}_\Sigma$ on-line from the relative frequencies of phonemes in the lexicon so far. That is, one occurrence of the phoneme is counted for each word type it appears in, not each word token, since we are interested in the probabilities of phoneme strings in the lexicon. The probability of # is estimated in the same way, except that there is exactly one # in each word type.

### 2.3.5 Distributions on the positive integers (model steps 1 and 3)

Equation (18) contains distributions on the positive integers in two places. The first is the

distribution on the total frequency of each word and the total number of utterance boundary markers (which, for better or for worse, have been given the same distribution). The second is the distribution on the total number of word types. We have no idea in advance that any particular positive integer is more likely than any other, so an ignorant prior makes the most sense. Since there is no uniform distribution on the positive integers, the best that can be done is a relatively flat distribution whose sum converges to one. In the experiments presented below, we opted for algebraic and computational simplicity by using:[5]

(20) $$\Pr(i) \equiv \frac{6}{\pi^2} \cdot \frac{1}{i^2}$$

for all choices of positive integers. Substituting into (18) yields the following formula for the relative probability of a novel word:

(21) $$\mathrm{R}(\overline{w}_k \mid w_k \notin L_{k-1}) = \frac{6}{\pi^2} \cdot \frac{n_k}{k} \cdot \frac{\Pr_\sigma(W_{n_k})}{1 - \frac{n_k - 1}{n_k} \cdot \sum_{j=1}^{n_k} \Pr_\sigma(W_j)} \cdot \left( \frac{n_k - 1}{n_k} \right)^2$$

Likewise, substituting (20) into (14) yields the following formula for the relative probability of a familiar word:

(22) $$\mathrm{R}(\overline{w}_k \mid w_k \in L_{k-1}) = \frac{\mathrm{f}_k(\hat{k})}{k} \cdot \left( \frac{\mathrm{f}_k(\hat{k}) - 1}{\mathrm{f}_k(\hat{k})} \right)^2$$

## 2.4 Discussion of the model.

In the previous three subsections an abstract, language-independent probability model was proposed, recursive formulae were introduced for evaluating the prior probability of any word sequence, and one possible way of filling out the details of the abstract model was worked

out. In working out the details, an approximation was introduced to provide for efficient evaluation of lexicon probabilities. It was also noted that in the experiments the probability distribution on phonemes ($\mathbf{Pr}_{\Sigma}$) is estimated from the input. The approximation speeds computation and the estimation improves performance over a uniform distribution on phonemes, but neither assumption is essential. Setting aside these details, equations (21) and (22) can be used, along with equation (5) to compute the exact prior probability (according to the model) of any given sequence of words. Even with the assumptions mentioned above, the probability of a word sequence is computed without estimating a probability distribution on words.

A related property of the abstract model is that the steps do not divide naturally into those that generate a stable grammar and those that generate a sample given the grammar. Specifically, outputs are not generated by sampling repeatedly from a stable distribution on words in the lexicon, which would qualify as a grammar. The model itself must be viewed as a single "universal grammar" and the corpus as a sample consisting of just one event.

According to the probability model, the values in equations (21) and (22) are not probabilities; they are merely ratios of probabilities of two unrelated corpus-generation events. However, it is interesting to compare them to what one would expect for the conditional probability of the *k*th word, given the first *k*-1 words. The first term of (22) is, sensibly enough, the relative frequency of the *k*th word in the corpus so far. However, this relative frequency counts the *k*th occurrence; it is not the normal relative frequency, computed only in terms of those occurrences whose presence in the input is guaranteed in the conditional probability conception. Since (22) is derived from marginal probabilities of whole corpora, the *k*th occurrence in the corpus whose probability is being evaluated is no different from any other occurrence. This has the effect of adding one to both the numerator and the denominator of the normal relative frequency. The result looks something like add-one smoothing, a technique that

is sometimes used in estimating the probabilities of words from their relative frequencies (Gale & Church, 1994; Witten & Bell, 1991). In the usual treatment of words as independent events, this kind of smoothing appears to be a correction to a probability model that is fundamentally unsuited to natural language vocabularies.[6] Under the current probability model, where an entire corpus is a single event in the probability space, "smoothing" falls out of the model.

The second term of (22) is zero for words with no previous occurrences — this makes sense, since (22) is an equation for *familiar* words. For words with one previous occurrence this term is one fourth, and it approaches one rapidly as the frequency of the word increases. This convergence to unity is not surprising from the perspective of words as independently sampled events from an unknown probability distribution; the observed relative frequency of a word becomes an increasingly accurate estimate of the true probability as the sample-size for that word gets larger. The observed relative frequencies of words that have occurred only once will tend to overestimate their true relative frequency (e.g., Church & Gale, 1991). To see this, consider the distribution of waiting-times for the first occurrence of a given word. This distribution will be the same for all words with the same true relative frequency. But the actual observed waiting times will be greater for some such words than for others. The sample of words that have actually been observed once in a finite corpus is not an unbiased sample; it favors words that, by chance, have waiting times at the low end of the distribution for words of their relative frequency.

Turning now to (21), the first term is a normalizing constant. The second term could be decomposed into $n_k$ and $1/k$ and the latter could be interpreted as a smoothed relative frequency. However, it may be more enlightening to think of $n_k/k$ as the type-token ratio — the average of the observed relative frequencies of all words that have occurred so far. A large type-token ratio suggests that words have not been repeated very often in the corpus so far, and therefore novel

words have occurred frequently. Conversely, a small type-token ratio suggests that words have been repeated frequently in the corpus so far, and therefore novel words have occurred relatively rarely. To the extent that the past is any predictor of the future, it makes sense to assign a higher probability to novel words in the future when novel words have occurred relatively more often in the past. The third term, which represents the probability that the particular novel word would be chosen at random during generation of the lexicon, is the dominant term. It has already been discussed at some length. Its most notable property is that it will tend to decrease rapidly with the length of the novel word under consideration, all other things being equal. The final term starts out at one-fourth for the first word type and rapidly approaches one as the number of word-types increases. It does not seem to have as natural an interpretation as the other terms, but it can be thought of in a somewhat similar way to the type-token ratio: The more often novel words have been observed in the past, the less reluctant one should be to posit them.

## 3 Optimization algorithms

Equations (5), (21), and (22) provide the means to compute the prior probability (according to the model) of any given sequence of words. In principle, these equations can be used to segment any observed corpus by computing the prior probabilities of all possible segmentations of the corpus and returning the one whose probability is greatest. The use of prior probabilities is sound because the posterior probabilities of segmentations of the observed corpus are proportional to their priors. This follows from the fact that observed corpora are generated deterministically from one of their segmentations at step 5 of the model — that is, each word sequence can yield one and only one observed corpus. However, segmentation by exhaustive search is not computationally tractable. In a corpus of $n$ phonemes with $m$ utterances, there are $2^{n-m}$ possible

segmentations — there can be a word boundary or not between each pair of adjacent phonemes, except that there is always a word boundary where there is an utterance boundary.

### 3.1 Incremental Search

This subsection presents an algorithm, Incremental Search, that is more computationally tractable than exhaustive search and provides a more plausible model of segmentation by humans. Incremental Search attempts to find the most probable segmentation of the entire corpus, exactly as exhaustive search does, the only difference being that Incremental Search does not evaluate the probability of every single segmentation. Instead, it evaluates the probabilities of segmentations of successively longer prefixes of the observed corpus, adding one utterance at a time. It searches for a local maximum in the prior probability of the segmentation of the entire prefix corpus by evaluating the probabilities of all segmentations that can be constructed as follows: Append some segmentation of the last utterance to the most probable segmentation found for the corpus of all previous utterances. For example, suppose that the most probable segmentation found for the corpus consisting of all previous utterances is $\overline{w}_m$ . Incremental Search evaluates the probabilities of all word sequences $\overline{w}_{m+p}$ , where $w_{m+1}, \ldots, w_{m+p}$ is a possible segmentation of the current utterance. The probability of each sequence is evaluated by the formula:

(23) $$\mathrm{Pr}(\overline{w}_{m+p}) = \mathrm{Pr}(\overline{w}_m)\prod_{i=1}^{p} \mathrm{R}(\overline{w}_{m+i}),$$

which follows directly from equation (4). Since $\mathrm{Pr}(\overline{w}_m)$ is fixed and in fact was computed when the previous utterance was processed, $\mathrm{Pr}(\overline{w}_{m+p})$ is obtained by computing and multiplying only *p* relative probabilities.

From a cognitive perspective, we know that humans segment each utterance they hear without waiting until the corpus of all utterances they will ever hear becomes available. Incremental Search, unlike exhaustive search, also has this property.

From a theoretical perspective, it is important to emphasize that although Incremental Search commits to segmentations one utterance at a time, it does so by optimizing the prior probabilities of segmentations of the entire corpus of all utterances processed so far. As discussed in the previous section, these computations do not require estimating a probability distribution on words. Indeed these computations can be exact as far as the abstract model goes, although exact computations may not be the best choice for a practical implementation.

The number of segmentations Incremental Search evaluates for each utterance is the binary exponential of the number of phonemes in the utterance minus one. To segment a corpus with $q$ utterances of lengths $l_1 \ldots l_q$ Incremental Search evaluates $\sum_{i=1}^{q} 2^{l_i - 1}$ segmentations, substantially fewer than the $2^{\sum_{i=1}^{q} l_i - 1}$ segmentations evaluated in the course of exhaustive search. Using current computers it is probably feasible to search through all possible segmentations of all but the rare, extremely long utterances. However, there is still room to improve the algorithm.

### 3.2 Dynamic programming

If we are willing to accept some approximations to equations (21) and (22) then a dynamic programming (i.e., Viterbi) type algorithm can be used to find the optimal segmentation of each utterance without evaluating all possible segmentations. The approximation is to use values of $L_k$, $k$, $n_k$, and $\mathbf{f}_k$ as of the end of the previous utterance, ignoring whatever changes may come about as a result of the current utterance. If $\overline{w}_m$ is the most probable segmentation found for the corpus

consisting of all previous utterances and $w_{m+1} \ldots w_{m+p}$ is a possible segmentation of the current utterance, the approximation is equivalent to using $R(\overline{w}_m w_{m+i})$ instead of $R(\overline{w}_m w_{m+1} \ldots w_{m+i})$, for $1 \le i \le p$. Under this assumption, (23) can be approximated by:

(24) $$\Pr(\overline{w}_{m+p}) \approx \Pr(\overline{w}_m) \cdot \prod_{i=1}^{p} R(\overline{w}_m w_{m+i})$$

This seems reasonable when $m \gg p$, so it can be expected that $L_{m+i} \approx L_m$, $m+i \approx m$, $n_{m+i} \approx n_m$, and $\mathbf{f}_{m+i} \approx \mathbf{f}_m$, for $1 \le i \le p$. When both the first and second occurrences of a novel word are in the same sentence this results in incorrectly assigning the novel-word probability to both occurrences. However, this is expected to be rare after the first few sentences.

The segmentation of the current utterance that maximizes (24) can be found by the algorithm shown in Figure 2, where *utterance* is the current utterance, $\overline{w}_m$ is the segmentation chosen for all previous utterances, and $R$ is a function that computes $R(\overline{w}_m w_{m+i})$ for any $\overline{w}_m$ and $w_{m+i}$.

**Insert Figure 2 about here.**

When $R$ is computed according to equations (21) and (22) this algorithm will be called MBDP-1.[7] *MBDP* stands for *model-based dynamic programming* and “1" signifies the hope that some components of the model will ultimately be improved, leading to revisions of the equations.

MBDP-1 processes each utterance in time proportional to the square of the number of phonemes in the utterance. Thus, the time required to segment a corpus with $q$ utterances of lengths $l_1 \ldots l_q$ is order $\sum_{i=1}^{q} l_i^2$, a substantial improvement over Incremental Search.

At the risk of redundancy, it is worth emphasizing once again that MBDP-1 evaluates word sequences by prior probability, without estimating a probability distribution on words.

## 4 Experiment

This section presents experiments comparing MBDP-1 to algorithms based on transitional probabilities (TP), mutual information (MI), simple recurrent nets (Elman), and probabilistic word grammars (Olivier). We also include a comparison to LZW, the compression scheme on which *compress* is based (see e.g. Hankerson, Harris, and Johnson, 1998). Finally, we compare all these algorithms to pseudo-random segmentation in which the correct number of word boundaries are inserted at random locations in the corpus. All the algorithms except the random one segment in a completely incremental, unsupervised fashion and start with no knowledge of the input language.

### 4.1 Method

#### 4.1.1 Implementations

*Transitional probabilities and mutual information.* These algorithms track the frequencies of all phonemes and phoneme bigrams in the portion of the corpus processed so far. At phoneme position $i$ they first update the frequencies for phoneme $i$, phoneme $i$+1, and the bigram spanning both. Next they compute the appropriate statistic (either transitional probability or mutual information) between $i$ and $i$+1, call it S[$i$]. Finally, they insert a boundary between

$i-1$ and $i$ if S[$i-1$] is less than both S[$i$] and S[$i-2$]. Word boundaries are inserted at utterance boundaries unconditionally, but otherwise each utterance boundary symbol is treated as a "phoneme" in the input.

*Elman's algorithm.* As in Elman (1990), we used a simple recurrent net (SRN) with 20 hidden units and 20 context units. Each of the 50 phonemes in our transcription system was represented as an arbitrary six-bit vector, so there were six input and six output nodes. The weights were bounded by one and minus one, the random initial weights were bounded by 0.1 and $-0.1$, the learning rate was 0.1, and the momentum was zero. At phoneme position $i$ the root-mean-squared difference between the output vector and the vector representing phoneme $i+1$ was computed. This statistic was used for segmentation in the same way as transitional probabilities and mutual information, except for the treatment of utterance boundaries. Word boundaries were inserted at utterance boundaries unconditionally as before, but following Elman, utterance boundaries were otherwise ignored, rather than being treated as input characters. Finally, the network was trained to predict phoneme $i+1$ by back-propagating the error. We also tried running this algorithm with 5, 10,...,100 hidden and context units, but since the performance was nearly identical for all values we do not report the results.

*Olivier's algorithm.* In our implementation of Olivier's algorithm the input was processed in blocks consisting of one utterance. A word-frequency table was initialized to contain all the phonemes of the input alphabet, each with frequency two. Before processing each utterance, relative frequencies were computed by dividing each word's frequency by the sum of the frequencies of all words in the table. Treating these relative frequencies as probabilities, a

maximum likelihood parse of the utterance was found. The frequencies of words in this parse were then added to their frequencies in the stored table. Finally, each pair of adjacent words in the maximum likelihood parse was concatenated and the frequency of the resulting string in the frequency table was incremented. Word boundaries were inserted unconditionally at utterance boundaries. Frequency-one words were deleted from the word grammar after every 500 utterances; varying this parameter did not improve segmentation performance.

*LZW.* LZW, the text-compression algorithm that forms the basis of *compress*, maintains a set of words, initialized to the source alphabet, and a pointer to the next unprocessed character in the input. Starting with the next unprocessed character, it segments out the longest string that matches a word in its current word set, outputs a compressed representation of the matching word, and advances the input pointer over the matching string. It then adds to its set of words the string just matched with the next character in sequence appended to the end. For example, after matching *ab* from the input *abcde*, it would advance the pointer to *c* and add *abc* to the word set. Each matching string is considered a word in the segmented text. In our implementation, matches were automatically terminated at utterance boundaries and new matches started afterward. When a match included the last character of an utterance, no new word was added to the lexicon.

*MBDP-1.* As described above MBDP-1 processes utterances one at a time. In our implementation equation (24) is used from the beginning, although the approximation it is based on is less accurate for the first few utterances than it is later on. After a segmentation is chosen for an utterance, data structures representing $n$, $L$, $k$, and f are updated to reflect that segmentation. The order of the words in the segmented utterances does not need to be stored to

compute R because equations (21) and (22) are derived from the assumption that all orders are equally probable.

*Random baseline*. We used a pseudo-random baseline segmentation to shed some light on whether the other six algorithms were useful for segmentation at all. This baseline segmentation was obtained by first counting the number of words in the correct, standard segmentation, then inserting that number of word boundaries at randomly chosen, distinct locations in the corpus.

### 4.1.2 Input

All algorithms were tested on the same corpus of phonemic transcripts of spontaneous child-directed English. Orthographic transcripts made by Bernstein-Ratner (1987) were taken from the CHILDES collection (MacWhinney & Snow, 1985) and transcribed phonemically. The speakers were nine mothers speaking freely to their children, whose ages averaged 18 months (range 13-21). In order to minimize the number of subjective judgments and the amount of labor required every word was transcribed the same way every time it occurred. Onomatopoeia (e.g., *bang*) and interjections (e.g., *uh* and *oh*) were removed for the following reasons: (1) They occur in isolation much more frequently than ordinary words, so they would have inflated performance scores; (2) their frequency is highly variable from speaker to speaker and transcriber to transcriber, so their presence would have increased the random variance in performance scores; and (3) there is no standard spelling or pronunciation for many of them, so we could not tell from the orthographic transcript what sound was actually uttered. The total corpus consisted of 9,790 utterances, 33,387 words, and 95,809 phonemes. The average of 3.4 words per utterance is

typical of spontaneous speech to young children. The average of 2.9 phonemes per word is not surprising for a transcription system like ours, where diphthongs, r-colored vowels (e.g. the "ar" of *bar*), and syllabic consonants (e.g., the second syllable of *bottle*) are each transcribed by a single symbol. These sounds are represented by two symbols in some transcription systems and sometimes more than two in English orthography. (For examples, see the sample output of MBDP-1 in Table 2). Before running the experiment all word boundaries were removed, but utterance boundaries were left intact.

#### 4.1.3 Procedure

Each segmentation algorithm was run on the corpus described above, as was the random baseline algorithm. The Elman algorithm is non-deterministic due to the random initial weights, so it was run 100 times, as was the random algorithm.

#### 4.1.4 Scoring

As an objective (though certainly imperfect) standard of correct segmentation we used the orthographic segmentation. The input was scored in two ways, one emphasizing the utility of the algorithm for segmentation and the other emphasizing its utility for discovering novel words. To compute the segmentation scores, we aligned each phoneme of the segmentation produced by each algorithm with the corresponding phoneme of the standard segmentation. Each word in the algorithmic segmentation was labeled a *true positive* if it lined up exactly with a word in the standard segmentation — that is, both boundaries matched. Each word in the algorithmic

segmentation which did not align exactly with a word in the standard segmentation was counted as a *false positive*. That is, a false positive was scored for each word in the algorithm's output unless *both* its boundaries aligned with *consecutive* boundaries in the standard. Each word in the standard segmentation which did not align exactly with a word in the algorithmic segmentation was counted as a *false negative*. That is, a false negative was scored for each word in the standard unless *both* its boundaries aligned with *consecutive* boundaries in the algorithm's output. Note that all measures assess exact matches of whole words, not matches of single boundaries. We then computed *precision* and *recall* as follows:

$$\text{precision} \equiv \frac{\text{true positives}}{\text{true positives} + \text{false positives}}, \qquad \text{recall} \equiv \frac{\text{true positives}}{\text{true positives} + \text{false negatives}}$$

These are identical to the measures that we have called *accuracy* and *completeness* in previous papers (e.g., Brent & Cartwright, 1996). In order to reveal how the amount of input processed affects the performance of each algorithm the corpus was divided into blocks consisting of 500 consecutive utterances. Segmentation precision and recall were scored separately for each block.

To get a better picture of how each algorithm performs a measure that we call *lexicon precision* was computed after each block of 500 utterances. After each block, each word type that the algorithm produced was labeled a true positive if that word type had occurred anywhere in the portion of the corpus processed so far; otherwise it is labeled a false positive. Because each distinct word type contributes only one point to the score for each block, lexicon precision is influenced less by performance on high frequency words and more by performance on low frequency words.

### 4.2 Results and Discussion

Sample output from MBDP-1 is shown in Table 2. The first five utterances are all treated as single, novel words. By 100 utterances segmentation is already fairly good, although utterance 104 is still under segmented. The only errors in utterances 1000-1004 are: *nose* is twice segmented into *no* and *se*, probably due to the high frequency of the word *no* and the fact that the sound *z* serves as a morphemic suffix (plural nouns and 3$^{rd}$ person singular verbs); *those* is segmented into *tho* and *se*, probably for similar reasons. The only error in the last five utterances of the corpus is the failure to segment *Ididn'tthinkitwould*. Overall, these examples suggest that most of the errors are either failures to segment a string of words or over segmentation at real or potential morpheme boundaries; there seem to be very few errors that split morphemes.

For all seven algorithms, the segmentation precision is shown in Figure 3, the segmentation recall in Figure 4, and the lexicon precision in Figure 5.

**Insert Figures 3, 4, and 5 about here.**

For the two non-deterministic algorithms the mean of 100 runs is shown. In both cases, there was very little variance from run to run — the standard error of the mean for every block of every score was less than .001 for both algorithms.

By every measure, MBDP-1 outperforms all the other algorithms. Except for Olivier's algorithm, the performance ranking of the algorithms is consistent across all three measures: MBDP-1 is better than MI, MI is better than TP, TP is better than Elman's algorithm, Elman's

algorithm is better than LZW, and LZW is better than the insertion of the correct number of word boundaries at random locations. In retrospect, it is not terribly surprising that the algorithms based on transitional probabilities and mutual information do not perform as well as the model-based algorithm. The length of words in natural languages is unbounded. In order to discover words, the ability to represent words of length greater than two phonemes would appear to be a distinct advantage. Further, one might expect that mutual information would be a better way of representing words of length two than transitional probability, since the mutual information measures the degree to which two phonemes tend to cooccur symmetrically. However, SRNs can represent statistics over strings of arbitrary length, so it is not clear why Elman's algorithm did not perform better. The only explanation seems to be that it does not learn the cooccurrence statistics of the corpus as well as transitional probabilities, perhaps because of its limited representational capacity. The lack of consistent improvement with corpus size beyond about 2,000 phonemes legislates against the notion that more input would help. The fact that all six of the unsupervised segmentation algorithms perform better than random segmentation suggests they are all doing something relevant to the segmentation task.

On the segmentation precision and recall scores there is a notable gap between the two worst performers, LZW and random, and the remaining algorithms. To see why LZW is so poorly suited to linguistic segmentation, note that a particular word *W* can only be segmented out once in a context where it is followed by a particular phoneme *p*; thereafter, *Wp* will be a longer match than *W*. For example, LZW can segment out only one of the instances of *to* followed by *the*, no matter how many there are in the input. More generally, each word type can be segmented

out at most as many times as there are phonemes in the source alphabet.

All the algorithms except for MBDP-1 show at least a slight decrease in the lexicon precision measure as more text is processed. The pattern of results for MI is particularly interesting: It gets moderately good and slightly improving scores on the segmentation measures, while it gets poor and decreasing scores on the lexicon. One possible explanation is as follows. Since MI stores the frequencies of all phonemes and pairs of adjacent phonemes, it can represent the common one- and two-phoneme function words explicitly, even though it has no explicit representation of longer words. By the lexicon score, which attends only to word types, MI gets full credit for segmenting out these words in the first few blocks. After that, it continues to posit new words, but with little success. However, since the short, common words account for a large fraction of the tokens in every part of a text, MI continues to do relatively well by the measures that give credit for each token.

For Olivier's algorithm our results were completely consistent with the results he reported: The probabilistic word grammar converged to a state where its words were far too long and the algorithm severely undersegmented. This is reflected in the difference between its segmentation precision and recall scores. Since a non-trivial proportion of the *utterances* in the input corpus consist of just a single word, moderately good segmentation precision could be achieved by treating each utterance as a single word. However, only a small proportion of the *word tokens* are in single-word utterances, so treating every utterance as a single word results in poor recall. Olivier's algorithm chooses words that are too long because it attempts to minimize the product of relative frequencies of words in an utterance without penalizing for the introduction of long new words into the grammar. Minimizing the product of relative

frequencies of words strongly favors minimizing the number of words — that is, maximizing the length of words — and nothing penalizes it. In our model the probability that a given word will be generated in the lexicon $(\Pr_{\sigma})$ decreases rapidly with the length of the word. This tendency balances the terms that favor minimizing the product of relative frequencies. We believe that this is the main factor accounting for the success of our algorithm. Indeed, any proper probability distribution on the set of all strings over a language will, in a certain sense, display a trend toward lower probabilities for longer words (Rissanen, 1989).

The de Marcken (1996) and the Christiansen, Allen, and Seidenberg (1998) algorithms are not directly comparable to ours, but it is worth mentioning the experimental results they report. De Marcken's algorithm is not comparable because it creates a hierarchical decomposition of the input rather than a segmentation. Nonetheless, de Marcken scores his decomposition by comparing it to the orthographic segmentation, as we have done here. He gives his algorithm credit for finding a word if any node in the hierarchical decomposition spans precisely that word. By this measure de Marcken reports a word recall of 90.5% on the million-word Brown corpus. However, only one in six internal nodes in the hierarchy spans a true word, yielding a word precision of less than 17%. De Marcken describes this as an advantage for his applications, but it is not good performance on the task addressed in this paper. Christiansen et al's algorithm is not directly comparable to MBDP-1 because it relies on input feature vectors that encode the phonological relationships among phonemes, while MBDP-1 treats phonemes as arbitrary, meaningless symbols. Christiansen et al. provide an encoding for the phonemes of one British dialect, but the performance of their algorithm on other dialects would depend on non-obvious decisions about how to encode phonemes as feature vectors. They report a segmentation recall of 44.9% and precision of 42.7% on a corpus of spontaneous child-directed British English with remarkably similar characteristics to our corpus.

It is also interesting to compare the performance of our algorithm to that of the batch segmentation procedure Brent and Cartwright (1996) used to test their proposed objective function. Brent & Cartwright's objective function, based on a minimum description length analysis, shares many significant features with the objective function derived here by a probabilistic analysis. Because the batch algorithm took time proportional to the cube of the number of phonemes in the corpus, it could not be tested on a corpus as large as the 10,000-utterance corpus used here. Brent and Cartwright reported average segmentation precision and recall for nine separate runs on nine 1350-phoneme (roughly 170-utterance) subsections of the corpus used here. The results were 41.3% precision and 47.3% recall, a little over half the performance levels reported here. This difference in performance results from the combined effects of four differences between the two algorithms:

1. the change from batch to incremental search, which makes it computationally feasible to process corpora of essentially unlimited size and hence to learn from a more representative sample of the language
2. the change from the "insert-two" heuristic search (which could be used in either a batch or an incremental algorithm) to dynamic programming
3. the change from generating the lexical entries according to a uniform distribution on phonemes to estimating a distribution on phonemes from the current hypothesized lexicon
4. differences in the objective function resulting from using the model-based approach rather than the minimum description length approach.

Finally, we were curious how MBDP-1 would perform in the ideal case in which it was

given a correctly segmented version of the corpus followed by an unsegmented version. This would have the effect that (a) all the target words in the unsegmented portion would be familiar words, and (b) at any point during the processing of the unsegmented portion, the relative frequencies of words in previous segmentations would be approximately correct. Even under these ideal conditions performance could be limited by genuine ambiguities where syntax or semantics is needed to decide among alternative ways of covering an utterance with actual words. In addition, differences between the true distribution on word sequences and the simple model on which MBDP-1 is based could lead to errors. However, as it turned out segmentation accuracy exceeded 98% under these conditions.

# 5 Conclusions

## 5.1 Summary of findings

This paper began with an abstract specification of a probability model to which more detailed models of phonology, word-order, and word frequency were fitted. Together with these detailed models, the abstract model yields a language-independent, prior probability distribution on all possible sequences of all possible words in any language over a given alphabet. The most important features of the model are:

1. the abstract specification, which makes it possible to improve parts of the model without affecting its fundamental structure
2. the generation of a lexicon and a complete corpus, regardless of length, as a single event in the probability space, which avoids a number of theoretical problems associated with

estimating probability distributions on vocabulary

3. the fact that there is a one-to-one correspondence between joint outcomes of the model’s nondeterministic steps (1-4) and segmentation hypotheses, which means that the prior probabilities of word sequences can be computed without summing over outcomes, at least as far as the abstract model goes

4. the fact that each segmentation hypothesis yields exactly one observable corpus via the deterministic step (5), which means that the posterior probabilities of segmentation hypotheses, given an observed utterance sequence, are proportional to their priors.

Two segmentation algorithms were obtained by adapting the probability distribution to two search methods: Incremental Search, which uses the probability model directly, and a dynamic programming search that requires an approximation. The latter combination was dubbed MBDP-1 and used in the experimental section. Both methods are based on the idea of finding, for any observed corpus, the segmentation of that corpus whose prior probability is greatest. Both progress through the corpus one utterance at a time, committing to the segmentation of that utterance before considering the next. The more efficient of the two algorithms, MBDP-1, processes each utterance in time proportional to the square of its length.

In an experimental comparison to other language-independent incremental segmentation algorithms, MBDP-1 robustly outperformed all the other algorithms tested. The most dramatic differences showed up in the lexicon precision measure, which moderates the influence of high frequency words.

## 5.2 Future work

There is room for improvement in all three of the components that fill out the details of the abstract probability model: word-order, phonology, and word frequency. For word-order, the current model assumes a uniform distribution on permutations of the word types. This is tantamount to ignoring all syntactic and semantic effects on the probabilities of various word sequences. A simple improvement would be to add a step to the model in which bigrams or trigrams of words are generated from the set of word types. Instead of choosing frequencies for each word independently, the model would choose frequencies for each word conditioned on each possible previous context. The permutations would then be chosen so as to respect these contextually conditioned frequencies. A more ambitious approach would be to add some kind of grammar — say, a stochastic context-free grammar — to be picked by the model according to some prior distribution on such grammars (perhaps along the lines of Stolcke, 1994).

For phonology, the current model generates word types for the lexicon by choosing phonemes independently of one another. This is tantamount to ignoring the effect of a language's phonology on the probabilities of various phoneme sequences being words of that language. Just as with word sequences, a relatively straightforward approach would be to use bigrams or trigrams of phonemes. Another approach would be to use template grammars to generate words in the lexicon (see Cartwright & Brent, 1997).

Improvements could also be made in modeling the relationship among the total frequencies of words, their lengths, and the number of utterances in the corpus. There have been several attempts to model word frequency distributions in corpora, length distributions, and the

joint length-frequency distribution in detail (Baayen, 1991; Mandelbrot, 1953; Miller, 1957; Zipf, 1935); incorporating the results of these studies would make the model more accurate.

While MBDP-1 offers substantial improvements over previous unsupervised segmentation algorithms, we look to improving it through the addition of better component models of phonology, word order, and word frequency.

**5.3 Implications for child language acquisition**

From the perspective of language acquisition theory MBDP-1 is consistent with the INCDROP model (Brent and Cartwright, 1996; Brent, 1996, 1997). Both models predict that learners will segment utterances in such a way as to:

1. Minimize the sum of the lengths of all novel words in the segmentation
2. Maximize the product of relative frequencies of all words in the segmentation, where the relative frequency of all novel words is taken to be a positive number less than the relative frequency of a word that has occurred only once before.

These predictions in turn imply that:

1. An utterance will be treated as a single, novel word if no substring of it is a familiar word.
2. Familiar words that do not overlap other familiar words in an utterance will tend to be segmented out, unless they are both short and rare.

Using adult subjects in artificial-language learning experiments, Dahan & Brent (1999) find evidence in favor of the latter two predictions. The models also make other, more detailed predictions, many of which have not yet been tested in behavioral experiments (see Brent, 1997;

Dahan & Brent, 1999).

From a cognitive modeling perspective, the main difference between MBDP-1 and our previous models is that MBDP-1 uses estimates of the frequency in the lexicon of each phoneme. The fact that this contributed to improved performance suggests that it would be useful for human language learners to do likewise. For example, it would be useful for children learning English to know that, while the initial phoneme of *the, this, that,* and *them* occurs very frequently in speech, it nonetheless occurs in very few distinct words. Thus, the hypothesized occurrence of a novel word like *thatman* should be assigned a lower probability than that of a novel word like *batman* because of the relatively low frequency of *th* in the child's developing lexicon. Thus, the model suggests that it might be worthwhile to investigate whether human learners do indeed (a) distinguish between the frequency of phonemes in the lexicon and their frequency in speech, and (b) use lexical frequency in assessing the plausibility of novel words.

## Acknowledgments

This work was supported in part by a grant from NIDCD (DC 03082) to Michael Brent, who is a member of the Center for Language and Speech Processing at Johns Hopkins University. Thanks to Anne Cutler and the Max Planck Institute for Psycholinguistics, where MB did part of the theoretical work presented here. Thanks to Adam Grove for commenting, “MDL is just poor-man’s Bayesian.” Thanks to Frantisek Kuminiak for pointing out the need to ensure that word types are not duplicated in the lexicon; to Mark Liberman for suggesting that we might find a way to use dynamic programming; to Colin Wilson and other students in my course for refusing to accept the minimum description length principle on faith; to Lidia Mangu for extensive and stimulating discussion of the mathematics; to Lidia and Anand Raman for careful reading of earlier versions of this paper and useful comments on them; to Anand for implementing and testing the LZW algorithm, and to Kip Lubliner for implementing and testing the Elman and Olivier algorithms.

## Appendix A: Approximation to the distribution on sets of pronunciations

This appendix presents the derivation of the approximation to the actual probability (16). Pulling out the $i = n_k$ term from the numerator of (16), we can rewrite the right-hand-side as:

$$\text{(25)} \qquad \Pr_\sigma(W_{n_k}) \frac{\sum_{p:\{1,\ldots,n_k\}} \left(1 - \sum_{j=1}^{n_k-1} \Pr_\sigma(W_{p(j)})\right)^{-1} \cdot \prod_{i=1}^{n_k-1} \left(1 - \sum_{j=1}^{i-1} \Pr_\sigma(W_{p(j)})\right)^{-1}}{\sum_{p:\{1,\ldots,n_k-1\}} \prod_{i=1}^{n_k-1} \left(1 - \sum_{j=1}^{i-1} \Pr_\sigma(W_{p(j)})\right)^{-1}}$$

We now use the approximation

$$\text{(26)} \qquad \left(1 - \sum_{j=1}^{n_k-1} \Pr_\sigma(W_{p(j)})\right)^{-1} \approx \left(1 - \frac{n_k - 1}{n_k} \cdot \sum_{j=1}^{n_k} \Pr_\sigma(W_{p(j)})\right)^{-1}$$

The sum on the left-hand-side includes the probabilities of all the $W_i$, $i = 1, \ldots, n_k$ except one, $W_{p(n_k)}$. The approximation is that instead of omitting precisely this probability, we omit the

probability of the average of all the $W_i$. This approximation is too great for some permutations and too small for others. Since the approximation is a sum over all the $W_i$ it is independent of the permutation and can be factored out, allowing us to rewrite (25) as:

$$(27)\qquad \frac{\Pr_\sigma(W_{n_k})}{1-\frac{n_k-1}{n_k}\cdot\sum_{j=1}^{n_k}\Pr_\sigma(W_j)}\cdot\frac{\sum\limits_{\mathrm{p}:\{1,\ldots,n_k\}}\prod\limits_{i=1}^{n_k-1}\left(1-\sum_{j=1}^{i-1}\Pr_\sigma(W_{\mathrm{p}(j)})\right)^{-1}}{\sum\limits_{\mathrm{p}:\{1,\ldots,n_k-1\}}\prod\limits_{i=1}^{n_k-1}\left(1-\sum_{j=1}^{i-1}\Pr_\sigma(W_{\mathrm{p}(j)})\right)^{-1}}$$

We now define:

$$(28)\qquad h_k \equiv \frac{\sum\limits_{\mathrm{p}:\{1,\ldots,n_k\}}\prod\limits_{i=1}^{n_k-1}\left(1-\sum_{j=1}^{i-1}\Pr_\sigma(W_{\mathrm{p}(j)})\right)^{-1}}{\sum\limits_{\mathrm{p}:\{1,\ldots,n_k-1\}}\prod\limits_{i=1}^{n_k-1}\left(1-\sum_{j=1}^{i-1}\Pr_\sigma(W_{\mathrm{p}(j)})\right)^{-1}}$$

Substituting into (27), we write:

$$(29)\qquad \frac{\Pr(L_k)}{\Pr(L_{k-1})}\approx\frac{\Pr_\sigma(W_{n_k})}{1-\frac{n_k-1}{n_k}\sum_{j=1}^{n_k}\Pr_\sigma(W_j)}\cdot h_k$$

and focus on understanding the range of values that $h_k$ may take on.

To this end, we now change the denominator of $h_k$ to an equivalent expression summed over all permutations of $\{1,\ldots,n_k\}$. Define a function **g** from permutations of $\{1,\ldots,n_k\}$ to permutations of $\{1,\ldots,n_k-1\}$ as follows:

$$(30)\qquad \mathrm{g}(\mathrm{p})(i)\equiv\begin{cases}\mathrm{p}(i) & \text{if } \mathrm{p}(i)<n_k\\ \mathrm{p}(n_k) & \text{if } \mathrm{p}(i)=n_k\end{cases}$$

This function maps $n_k$ permutations of $\{1,\ldots,n_k\}$ into each permutation of $\{1,\ldots,n_k-1\}$, so we can rewrite (28) as:

$$h_k = \frac{\sum\limits_{\mathrm{p}:\{1,\ldots,n_k\}}\prod\limits_{i=1}^{n_k-1}\left(1-\sum_{j=1}^{i-1}\Pr_\sigma(W_{\mathrm{p}(j)})\right)^{-1}}{\frac{1}{n_k}\sum\limits_{\mathrm{p}:\{1,\ldots,n_k\}}\prod\limits_{i=1}^{n_k-1}\left(1-\sum_{j=1}^{i-1}\Pr_\sigma(W_{\mathrm{g}(\mathrm{p})(j)})\right)^{-1}}$$

The effect of changing **p** to **g**(**p**) in the denominator is to replace $\Pr_\sigma(W_{n_k})$ with $\Pr_\sigma(W_{p(n_k)})$ whenever the former term is present in the inner sum. Thus, we can change back from **g**(**p**) to **p** if we make this change explicitly, rewriting the previous equation as:

$$h_k = n_k \cdot \frac{\sum_{p:\{1,\ldots,n_k\}} \prod_{i=1}^{n_k-1} \left(1 - \sum_{j=1}^{i-1} \Pr_\sigma(W_{p(j)})\right)^{-1}}{\sum_{p:\{1,\ldots,n_k\}} \prod_{i=1}^{p^{-1}(n_k)-1} \left(1 - \sum_{j=1}^{i-1} \Pr_\sigma(W_{p(j)})\right)^{-1} \prod_{i=p^{-1}(n_k)}^{n_k-1} \left(1 + \Pr_\sigma(W_{n_k}) - \Pr_\sigma(W_{p(n_k)}) - \sum_{j=1}^{i-1} \Pr_\sigma(W_{p(j)})\right)^{-1}}$$

The numerator and the denominator are now identical except for the replacements of $\Pr_\sigma(W_{n_k})$ with $\Pr_\sigma(W_{p(n_k)})$. In permutations where $p^{-1}(n_k) = 1$ the replacement occurs in every term of the product. In permutations where $p^{-1}(n_k) = n_k - 1$ the replacement occurs in only one term of the product. In permutations where $\Pr_\sigma(W_{n_k}) > \Pr_\sigma(W_{p(n_k)})$ the substitution makes the term for that permutation smaller, and this contributes to making $h_k$ as a whole larger. Conversely, in permutations where $\Pr_\sigma(W_{n_k}) < \Pr_\sigma(W_{p(n_k)})$ the substitution makes the term for that permutation larger, and this contributes to making $h_k$ as a whole smaller. Thus, if $\Pr_\sigma(W_{n_k})$ is large compared to the probabilities of earlier words, the net effect will be to increase $h_k$ and hence to increase the relative probability of $L_k$, and conversely. In the experiments reported in this paper, we use the approximation $h_k \approx n_k$ and hence

(31) $$\frac{\Pr(L_k|n_k)}{\Pr(L_{k-1}|n_k - 1)} \approx \frac{n_k \Pr_\sigma(W_{n_k})}{1 - \frac{n_k - 1}{n_k} \cdot \sum_{j=1}^{n_k} \Pr_\sigma(W_{p(j)})}$$

with the understanding that this expression is not quite as responsive to $\Pr_\sigma(W_{n_k})$ as it should be. Equation (31) is the approximation given in the main text as (17).

In the experiment reported here the sum in the denominator is 0.302 after segmenting all 10,000 utterances. This suggests that requiring each string selected to be unique has some impact, increasing the probability of novel words by a factor of almost 1.5. This impact is of the same order as all of the other terms in the relative probability of a novel word (21) except for $\Pr_{\sigma}(W_{n_k})$, the probability of picking that word's particular phoneme string independent of the rest of the lexicon. Further, a better phonological model than the one used in these experiments would assign greater probability to the words in the lexicon (and less to the others). This could greatly increase the sum in the denominator, giving this term a greater impact on the overall probability.

Table 1: An example of a possible output of the generative process using ordinary letters for the alphabet.

1. $n$=6

2. L={do#, the#, kb#, like#, see#, mbo#}.

| $W_1$ | $W_2$ | $W_3$ | $W_4$ | $W_5$ | $W_6$ | $W_0$ |
|---|---|---|---|---|---|---|
| do# | the# | kb# | like# | see# | mbo# | $ |

3.

| **f**(1) | **f**(2) | **f**(3) | **f**(4) | **f**(5) | **f**(6) | **f**(0) |
|---|---|---|---|---|---|---|
| 2 | 4 | 2 | 1 | 2 | 2 | 2 |

4. $m$=2+4+2+1+2+2+2=15

| **s**(1) | **s**(2) | **s**(3) | **s**(4) | **s**(5) | **s**(6) | **s**(7) | **s**(8) | **s**(9) | **s**(10) | **s**(11) | **s**(12) | **s**(13) | **s**(14) | **s**(15) |
|---|---|---|---|---|---|---|---|---|---|---|---|---|---|---|
| 1 | 3 | 5 | 2 | 6 | 0 | 5 | 2 | 2 | 0 | 1 | 3 | 4 | 2 | 6 |

$w_1=W_{\mathbf{s}(1)}=W_1$=do#  $w_2=W_{\mathbf{s}(2)}=W_3$=kb#  $w_3=W_{\mathbf{s}(3)}=W_5$=see#  $w_4=W_{\mathbf{s}(4)}=W_2$=the#  …

5. dokbseethembo$seethethe$dokblikethembo

Table 2: Sample output from MBDP-1 run on a corpus of phonemically transcribed, spontaneous child-directed speech.

| Utterance | Phonetic Output (actual form) | Orthographic Equivalent |
|---|---|---|
| 1 | yuwanttusiD6bUk | youwanttoseethebook |
| 2 | lUkD*z6b7wIThIzh&t | lookthere’saboywithhishat |
| 3 | &nd6dOgi | andadoggy |
| 4 | yuwanttulUk&tDIs | youwanttolookatthis |
| 5 | lUk&tDIs | lookatthis |
| 100 | h9 d&d& | hi dada |
| 101 | se h9 d&d& | say hi dada |
| 102 | hElo | hello |
| 103 | hElo d&d& | hello dada |
| 104 | Iz Itd&dianD6fon | is itdaddyonthephone |
| 1000 | 6 no z | a no se |
| 1001 | It Iz 6kQz no z | it is acow’s no se |
| 1002 | r9t D* Iz 6kQz no z | right there is acow’s no se |
| 1003 | gUd g3l | good girl |
| 1004 | 9 dont TINk yu no Eni 6v Do z TIN z | I don’t think you know any of tho se thing s |
| 9786 | D* | there |
| 9787 | nQ D6 d% Iz op~ | now the door is open |
| 9788 | yu k&n pUt hIm In h( | you can put him in here |
| 9789 | D* | there |
| 9790 | no 9dId~tTINkItwUd fIt iDR | no Ididn’tthinkitwould fit either |

Figure 1: An example of the correspondence between Huffman codes and words in Brent and Cartwright's self-delimiting representation scheme. The leaves of the code tree contain the codes. The words are listed in order so that the leftmost word corresponds to the leftmost code in the tree. For example, if the sequence of words were *the#kitty#like#do#you#see* then the code word for *the* would be 00, the code word for *kitty* would be 01, and so forth. The order of the words in the segmentation *do you see the kitty see the kitty do you like the kitty* would then be represented by 10111011100011110001101110100001. Note that both the code tree and the letters in the word list are further encoded as self-delimiting binary strings (see Brent & Cartwright, 1996; Quinlan & Rivest, 1989).

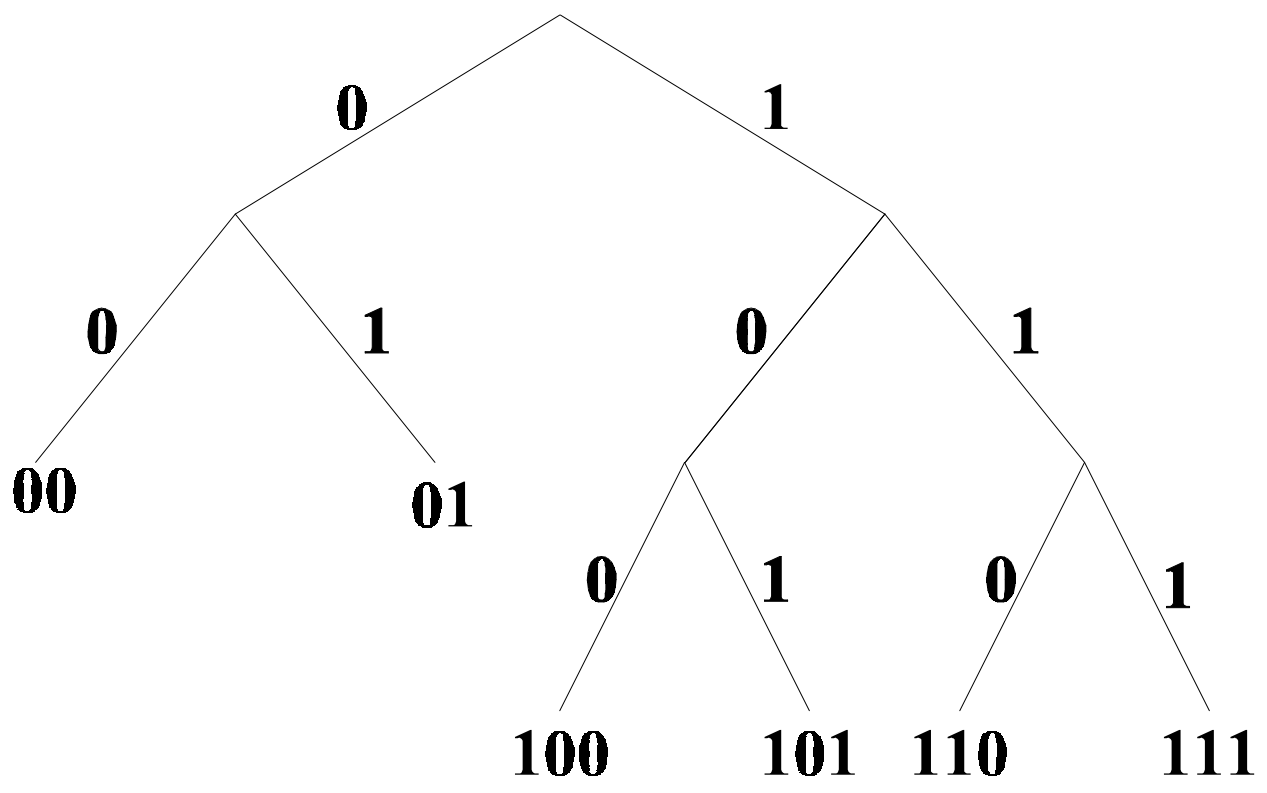

Figure 2: A dynamic programming algorithm for finding the segmentation of the current utterance that maximizes (24). The inputs are *utterance*, the current utterance, and $\overline{w}_m$ , the segmentation chosen for all previous utterances. *R* is a function that computes $R(\overline{w}_m w_{m+i})$ for any $\overline{w}_m$ and any hypothesized word $w_{m+i}$ according to equations (21) and (22). The indices *first-char* and *last-char* index arrays parallel to the input utterance.

```
MBDP-1(utterance, w̄_m )
for last-char = 0 to length(utterance)
        best-product [last-char] = R( w̄_m , substring(utterance, 0, last-char));
        best-start [last-char] = 0;
        // After this loop, best-start [last-char] points to the beginning of the optimal word
        // ending with last-char. Of course it may turn out that no word will end at last-char in
        // the optimal segmentation.
        for first-char = 1 to last-char
                word-score = R( w̄_m , substring(utterance, first-char, last-char));
                if word-score*best-product [first-char −1] > best-product [last-char];
                        best-product [last-char] = word-score * best-product [first-char − 1];
                        best-start [last-char] = first-char;
// Now work backward along the best path to insert actual word boundaries.
first-char = best-start [length(utterance)];
while first-char > 0
        insert-boundary(first-char);
        first-char = best-start [first-char - 1];
```

Figure 3: Segmentation precision of seven algorithms scored on successive 500-utterance blocks of phonemically-transcribed, spontaneous child-directed speech.

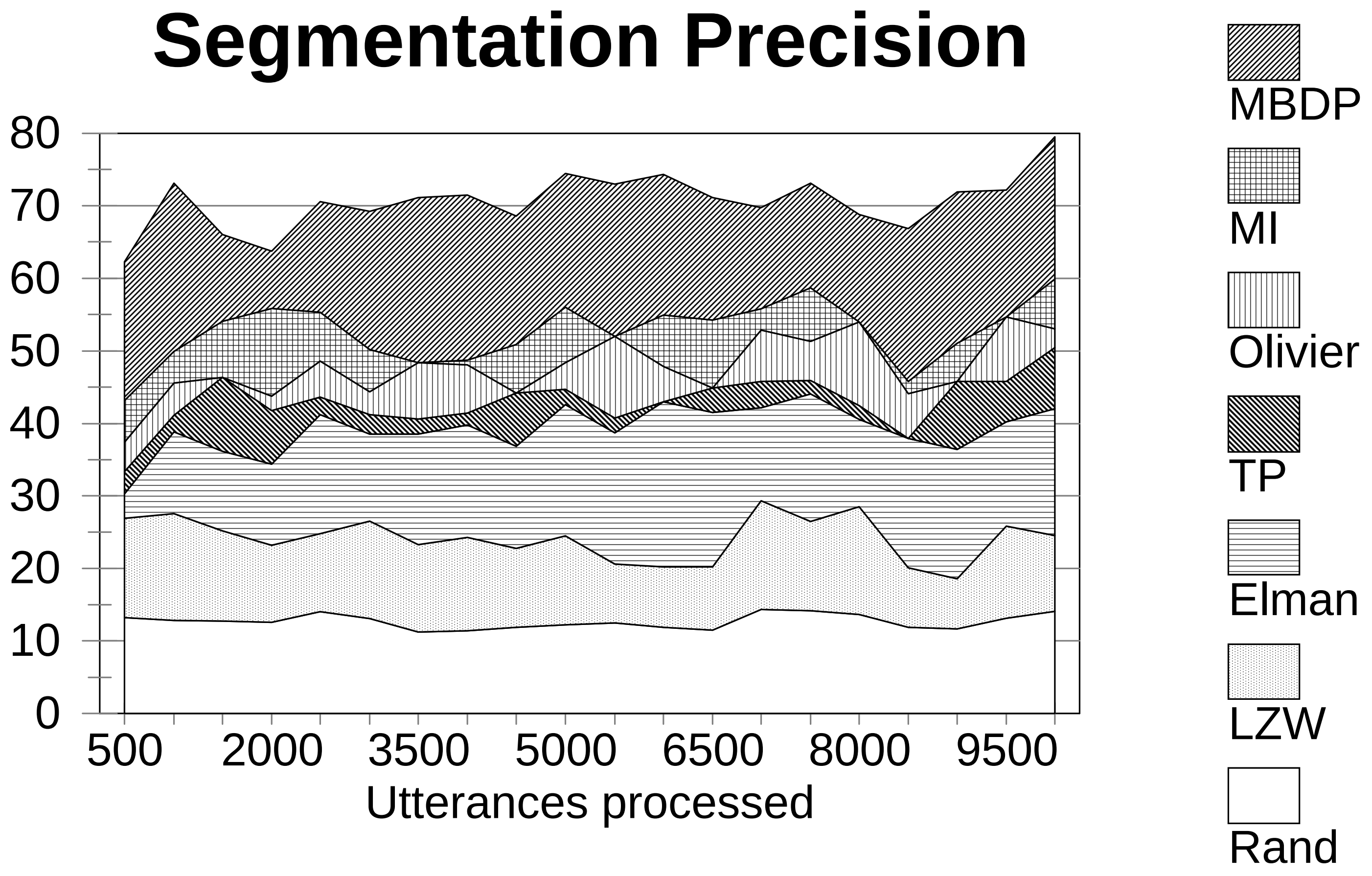

<u>Figure 4:</u> Segmentation recall of seven algorithms scored on successive 500-utterance blocks of phonemically-transcribed, spontaneous child-directed speech.

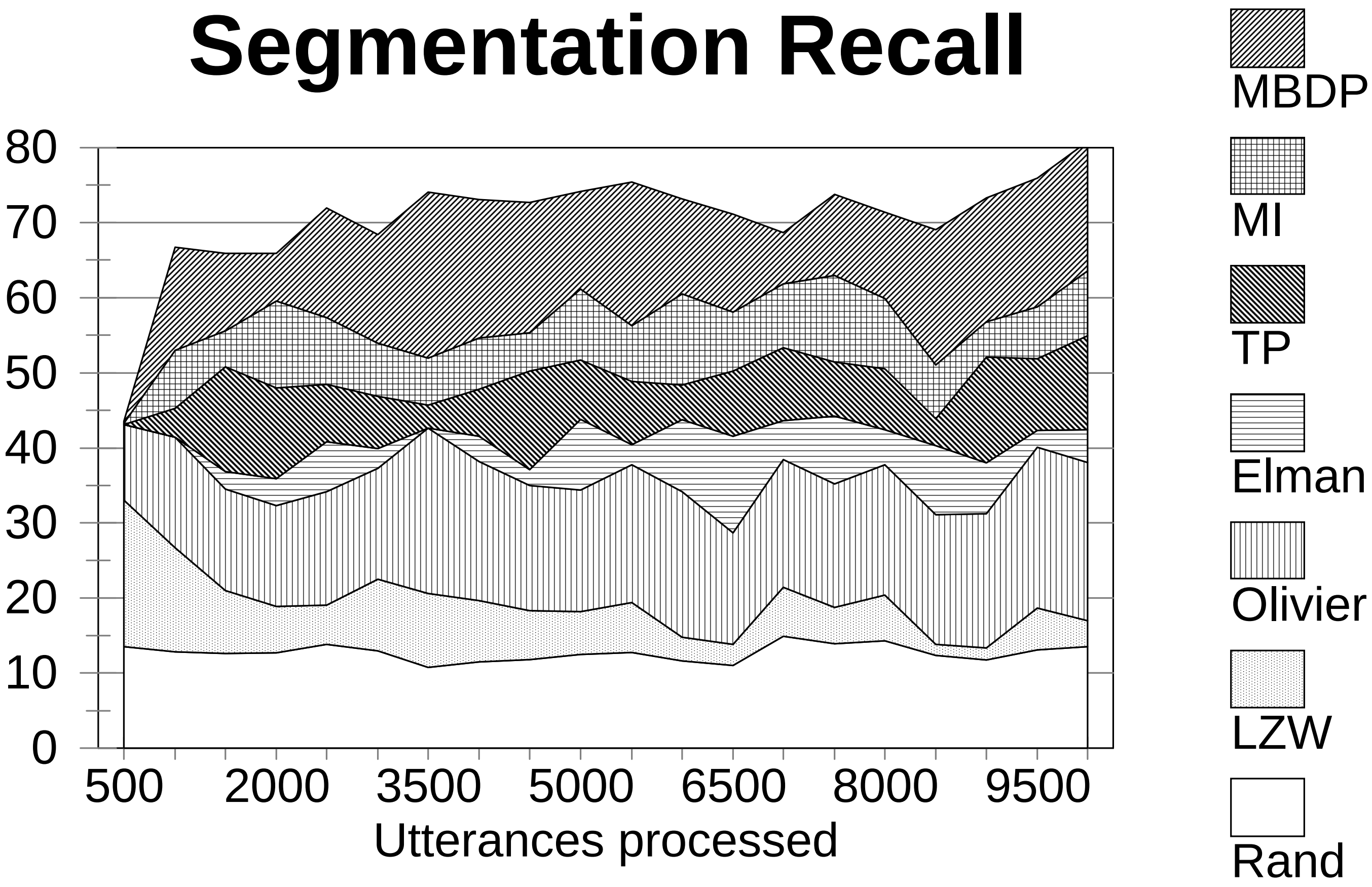

Figure 5: Lexicon precision of seven algorithms scored on successive 500-utterance blocks of phonemically-transcribed, spontaneous child-directed speech.

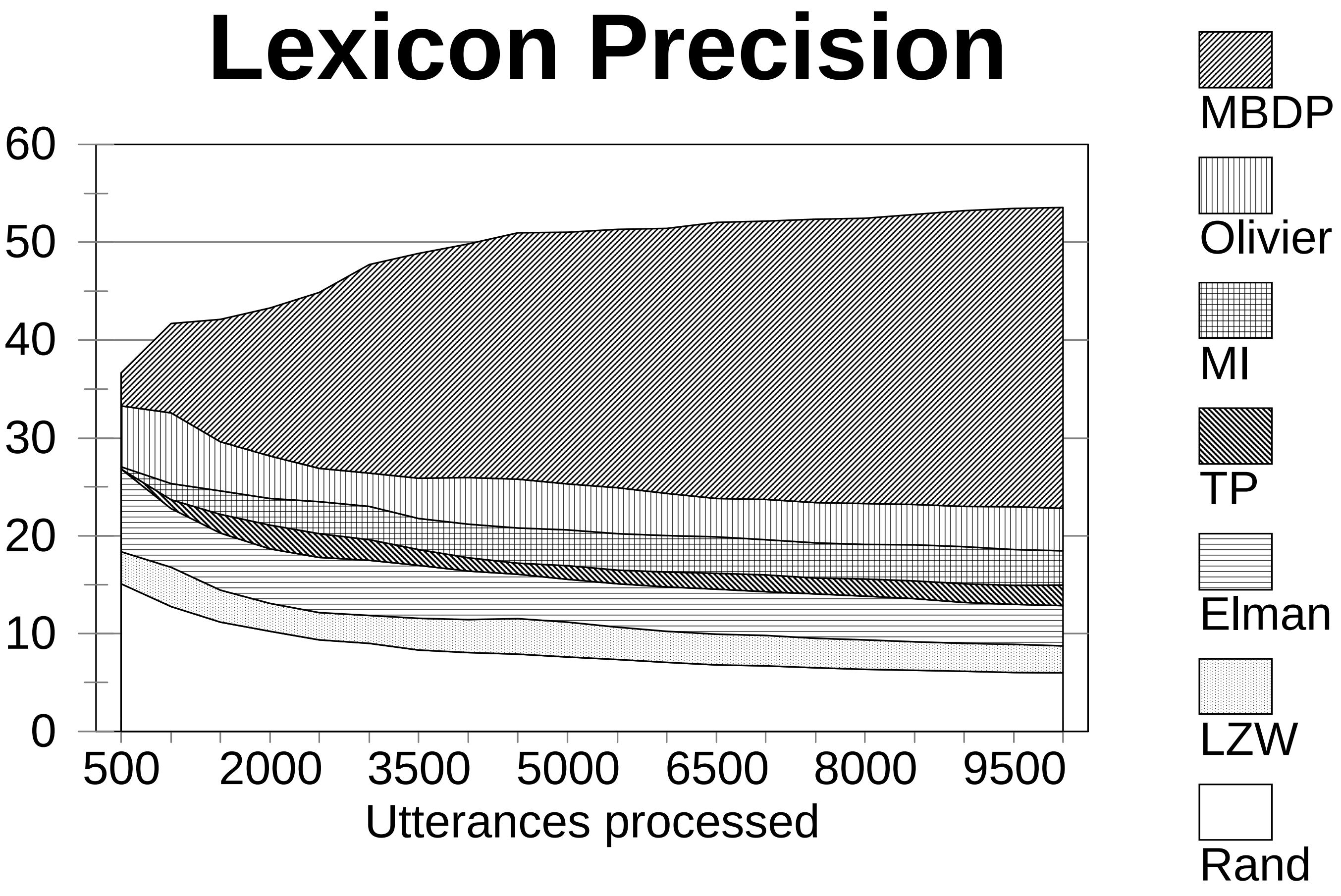

## Notes

1.*Phonemes* are symbols representing the basic sounds that serve to distinguish one word from another in a language. For example, the *b* sound of *cab* or the *sh* sound of *ship*.

2.Brent and Cartwright (1996) proposed an objective function that children might use for segmentation, but did not specify a search procedure for the optimization. Brent and Cartwright investigated their objective function using a global optimization algorithm, but this was not proposed as an algorithm children might use.

3.A morpheme is an atomic unit of meaning or syntactic function, including root words like *toy*, inflectional suffixes like -*s*, and derivational suffixes like -*ness*.

4.This ratio looks like a conditional probability, $\Pr(\overline{w}_k | \overline{w}_{k-1})$. Indeed, that notation would be convenient and algebraically correct. However, the semantics of the model are not consistent with a conditional probability interpretation. The sequence $w_1, \ldots, w_k$ is not a conjunction of events from the probability space but rather a single event that is determined by the joint outcomes of steps 1-4 above. Thus, $w_1, \ldots, w_{k-1}$ and $w_1, \ldots, w_k$ are actually distinct, mutually exclusive events from the probability space. The only reasons for taking their ratios are to simplify algebra and facilitate computation.

5.A reasonable alternative would be Rissanen's so-called universal prior (e.g, Rissanen 1989), which is the limit of a sequence of increasingly flat, monotonically decreasing distributions on the positive integers. Although the universal prior is flatter ("more ignorant") than the distribution we used, it is algebraically and computationally more complex.

6.Add-one smoothing can be derived rigorously under the assumption that the number of types is a known, fixed integer. When occurrences of words are treated as independent events this predicts that, as the size of the sample grows without bound, the frequency of the least frequent word will also grow without bound. This appears to be false for natural language vocabularies; large corpora are not sampled from a fixed vocabulary but reflect the continual generation of new vocabulary. When a corpus is viewed as a single event there is no such prediction — sample size is always one.

7.At the level of detail appropriate for cognitive modeling, MBDP-1 can be viewed as an implementation of the model that Brent (1997) and Dahan and Brent (1999) refer to as INCDROP.